\documentclass[twoside,11pt]{article}
\usepackage{jair, theapa, rawfonts}
\usepackage{standalone}
\usepackage{booktabs}

\jairheading{79}{2024}{1-40}{07/2023}{01/2024}
\ShortHeadings{Undesirable Biases in NLP: Addressing Challenges of Measurement}{Van der Wal, Bachmann, Leidinger, Van Maanen, Zuidema, \& Schulz}
\firstpageno{1}

\usepackage{graphicx}
\usepackage[utf8]{inputenc}

\DeclareUnicodeCharacter{1F99C}{\raisebox{-5pt}{\includegraphics[scale=0.08]{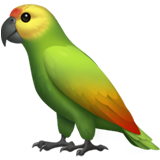}}}

\usepackage{multirow}

\usepackage{url}

\DeclareRobustCommand{\VAN}[3]{#2} %

\usepackage[most]{tcolorbox}
\newtcolorbox[auto counter,
number freestyle={\noexpand\Roman{\tcbcounter}},
]{application}[2][]{%
colframe=black,fonttitle=\bfseries,
colbacktitle=black,
coltitle=white,
enhanced,
title=Application~\thetcbcounter: #2,#1}

\begin{document}

\title{Undesirable Biases in NLP: \\Addressing Challenges of Measurement}

\author{\name Oskar van der Wal \email o.d.vanderwal@uva.nl\\
  \addr Institute for Logic, Language and Computation, University of Amsterdam
  \AND
  \name Dominik Bachmann
  \email d.bachmann@uva.nl \\
  \addr Institute for Logic, Language and Computation, University of Amsterdam\\
  Department of Experimental Psychology, Utrecht University
  \AND
  \name Alina Leidinger \email a.j.leidinger@uva.nl \\
  \addr Institute for Logic, Language and Computation, University of Amsterdam
  \AND
  \name Leendert van Maanen \email l.vanmaanen@uu.nl \\
  \addr Department of Experimental Psychology, Utrecht University
  \AND
  \name Willem Zuidema \email w.h.zuidema@uva.nl \\
  \name Katrin Schulz \email k.schulz@uva.nl \\
  \addr Institute for Logic, Language and Computation, University of Amsterdam
      }

\maketitle

\begin{abstract}
As Large Language Models and Natural Language Processing (NLP) technology rapidly develop and spread into daily life, it becomes crucial to anticipate how their use could harm people.
One problem that has received a lot of attention in recent years is that this technology has displayed harmful biases, from generating derogatory stereotypes to producing disparate outcomes for different social groups. %
Although a lot of effort has been invested in assessing and mitigating these biases, our methods of measuring the biases of NLP models have serious problems and it is often unclear what they actually measure.
In this paper, we provide an interdisciplinary approach to discussing the issue of NLP model bias by adopting the lens of psychometrics
--- a field specialized in the measurement of concepts like bias that are not directly observable.
In particular, we will explore two central notions from psychometrics, the \emph{construct validity} and the \emph{reliability} of measurement tools, and discuss how they can be applied in the context of measuring model bias. Our goal is to provide NLP practitioners with methodological tools for designing better bias measures, and to inspire them more generally to explore tools from psychometrics when working on bias measurement tools.
\end{abstract}

\section{Introduction}

In the last decade, technology for Natural Language Processing (NLP) has seen a very steep line of improvement. As a consequence, companies, governments and other institutions choose to employ this technology in more and more applications that directly impact the lives of ordinary citizens: Online customers are offered information on products that are automatically translated \shortcite<e.g.,>{way2018quality}, jobseekers are matched to vacancies based on automatic parsing of their resumes \shortcite<e.g.,>{montuschi2013job}, conversations with customer services, help desks and emergency services are automatically transcribed and analyzed to improve service \shortcite<e.g.,>{verma_natural_2011}, millions of medical and legal texts are automatically searched to find relevant passages, at times supporting decisions that may literally be matters of life and death \shortcite<e.g.,>{wang2018clinical,zhong2020does}. Most likely, NLP technology will soon be even more powerful and omnipresent, in light of recent developments, with larger datasets, bigger architectures, wider access to such models and the development of multipurpose models that can be applied to a multitude of different tasks~\shortcite{bommasani_opportunities_2022b}.

NLP technology, however, is far from error-free. In recent years various examples of NLP applications were brought to the public attention that behaved in ways that are harmful for certain individuals or groups: Systems for matching vacancies may unintentionally disadvantage ethnic minorities or people with disabilities \shortcite{hutchinson-etal-2020-social}, machine translation systems have been found to translate gender-neutral terms to the majority gender, which can amplify existing gender biases \shortcite{stanovsky2019evaluating}, speech recognition systems have difficulties to correctly recognize the voices of speakers of minority dialects \shortcite{zhang2022mitigating}, and, more generally, the biases and misinformation that generative models propagate can distort people's worldviews in unprecedented ways \shortcite{kidd2023ai}.

To combat these effects of language technology on society, detecting undesirable biases in Large Language Models and other NLP systems, and finding ways to mitigate them, has emerged as a new and active domain of NLP research. 
However, both detection and mitigation face problems. 
One of these challenges is that we lack sound tools to measure bias that is present in NLP systems. 
While there had been a lot of excitement about some early methods used to make bias in such systems visible \shortcite<e.g.,>{bolukbasi_man_2016,caliskanSemanticsDerivedAutomatically2017}, more recent work has shown that these (as well as newer) methods are problematic.
Many problems have been pointed out for how bias is defined and operationalized~\shortcite<see e.g.,>{blodgett_language_2020,blodgett_stereotyping_2021,dev_measures_2022,ethayarajh_understanding_2019,gonen_lipstick_2019,nissim2020fair,talat_you_2022}. There are also concrete issues with the measurement results. For instance, for some of the currently-used bias measures little to no evidence has been found that they correlate with other bias measures or with downstream harms~\shortcite<e.g.,>{cao_intrinsic_2022,delobelle_measuring_2022,goldfarb-tarrant_intrinsic_2021}.

If researchers cannot guarantee that bias measures for current NLP models work properly, it becomes difficult to make meaningful progress in understanding the scale of the problem and in designing mitigation strategies for the potential harms that may result from biased models. 
Using poor quality bias measurement tools could also give us a false sense of security when these measures show no or little bias.
Good design of such bias measures is thus critical. 
Consequently, we need ways to evaluate the quality of bias measures.

Helpful in that endevor could be knowledge from psychometrics. 
Psychometrics is the subfield of psychology concerned with the measurement of properties of human minds (e.g., intelligence or self-control) that cannot be directly observed. Treating bias as exactly such an unobservable \emph{construct} offers NLP new perspectives on conceptual problems concerning the notion of bias, and provides access to a rich set of tools developed in psychometrics for measuring such constructs. In this paper, we explore whether a psychometric view on bias in NLP technologies might offer a way to significantly improve the quality of bias measures.

Specifically, we focus on two concepts from psychometrics that are useful in the context of measuring notions as ambiguous as bias: construct validity and reliability. These concepts help us understand (a) what we measure, and how it relates to what we want to measure, and (b) how much we can trust the information provided by a specific bias measure. 
After introducing these concepts, we will explore how they can be interpreted and applied in the context of measuring bias in NLP. Our goal is to inspire and encourage other NLP practitioners to apply these concepts when developing and evaluating methods to measure bias in NLP technology.

We will start by discussing psychometric's distinction between constructs and their operationalizations, and explain why it is useful to view model bias in this framework (Section \ref{sec:unobservable_concepts}).
We then discuss reliability (Section \ref{sec:assess_reliability}) and construct validity (Section \ref{sec:assess_validity}), and the use of these concepts when evaluating bias measures in an NLP context.
Section \ref{sec:generalizing-bias} brings these concepts together in guiding questions for designing proper bias measures.

This is not the first paper proposing that AI researchers should utilize tools from psychometrics. For instance, \shortciteauthor{jacobs_measurement_2021} \citeyear{jacobs_measurement_2021} argue for applying psychometrics to study algorithmic fairness 
--- a discussion we now extend to NLP bias measures. In section \ref{sec:related-works} we will consequently position our paper in the literature and compare our contributions to those of related works \shortcite[i.a.]{bommasani2022trustworthy,du_assessing_2021,jacobs_measurement_2021}.

\section{Measuring Bias as an Unobservable Concept}
\label{sec:unobservable_concepts}

As pointed out in the introduction, measuring and mitigating bias in NLP systems has received a lot of attention in the last couple of years, and a broad range of tools has been developed to measure bias in NLP systems. Section~\ref{subsec:intro_bias_measures} provides a quick overview of these tools for readers not working in this area. 

Still, the existing bias measures face many problems \shortcite<see e.g.,>[as well as Section \ref{subsec:why_measure_bias}]{balayn2021beyond,blodgett_language_2020,cheng_socially_2021,kiritchenko_confronting_2021,orgad2022choose,talat_you_2022,weinberg_rethinking_2022}. Some authors have argued that these problems are at least partly a consequence of there not being a clear conceptualisation for or consensus about what we mean when talking about ``bias'' in NLP \shortcite{blodgett_language_2020,dev_measures_2022,talat_you_2022}. In light of this, one could argue that the field should instead look at better defined concepts, such as stereotyping or (downstream) harms. In Section~\ref{subsec:why_measure_bias}, we will explain why we still consider measuring model bias valuable. Additionally, we will argue that lack of agreement on a concept is not a problem in itself, as long as researchers are transparent about their assumptions. To this end, we will propose building on work from psychometrics and treating bias as a \emph{construct}. This allows NLP researchers to communicate their assumptions more precisely (see Section \ref{subsec:construct_vs_operationalization}) and test the quality of their bias measures. 

While one has to approach the translation of psychometric methodology and language with care and transparency, as will be explained in Section \ref{subsec:translation_step}, we think that there is a lot of promise in connecting NLP research to work in psychometrics. The rest of Section \ref{sec:unobservable_concepts} is hence dedicated to the introduction of some key concepts from psychometrics that we consider particularly useful for measuring bias in NLP: the difference between a construct and its operationalization in Section \ref{subsec:construct_vs_operationalization} and the notions of validity and reliability in Section \ref{subsec:construct_validity_reliability}.

\subsection{A Brief Introduction to Bias Measures in NLP}
\label{subsec:intro_bias_measures}

In the last decade, various methods have been developed to assess biases in NLP systems. Serving as background knowledge for our later discussions, we will here provide a short overview of some popular methods (see also Table~\ref{tab:example-bias-measures}).

Most early bias measures \shortcite<e.g.,>{bolukbasi_man_2016,caliskanSemanticsDerivedAutomatically2017} were designed for static word embeddings (such as word2vec, GloVe, or the input embeddings used in more recent language models), and often involved lists of word pairs that illustrate a particular semantic contrast (for instance, word lists contrasting gender, like \{\textit{man, he, boy}\} versus \{\textit{woman, she, girl}\}). Bias is then measured by looking at the similarity (or another geometric relationship in the embeddings space) of other words, for instance, terms referring to professions, to either the words in the male or the female list. For example, if an occupation word like ``manager'' shows stronger similarity with male words than with female words, this stereotype-consistent association is taken to show that the representation of this word is biased.
For an example, see \emph{Bias Direction} in Table \ref{tab:example-bias-measures}.

More recently, researchers have focused more on \emph{benchmark data sets} (or \emph{challenge sets}): data sets aimed at uncovering undesirable biases or stereotypes in language models, often designed as pairs or triples of sentences \shortcite<e.g.,>{nangia_crows-pairs_2020,webster2020measuring,zhao_gender_2018}. 
The general idea here is to compare the performance (or assigned probabilities) of the model on bias-consistent expressions with its performance on bias-inconsistent expressions. If the language model performs better on stereotype-consistent than on stereotype-inconsistent sentences, that is taken to indicate bias. 
For examples, see \emph{CrowS-Pairs}, \emph{STS-B for genders}, and \emph{WinoBias} in Table \ref{tab:example-bias-measures}.

Another paradigm that is gaining traction in evaluating language models, is the use of prompting. In prompting, large language models like the one underlying OpenAI's ChatGPT\footnote{\url{https://chat.openai.com}} are given a textual input along with an instruction in natural language that explains the task that has to be performed\footnote{For instance, one provides the model with a text describing a natural language inference (\texttt{All penguins have wings. Peter is a penguin. Therefore, Peter has wings}) and asks the model to evaluate whether the inference is logically valid.} \shortcite{liu_pre-train_2023}.
Prompting can also be used to assess the model's biases, for example, by providing the model with an item from a bias benchmark and asking it whether the item contains a stereotype~\shortcite{schick2021self} or which description of a person or social group (one being stereotypical, the other not) is more likely~\shortcite{scao_bloom_2022b}.
Many large language models are only accessible through a prompting API---the internals hidden away by the companies who created them. This is unfortunate because evaluation by prompting also has limitations \shortcite<including the fact that there is no guarantee that these models remain the same over time; see also e.g.,>[]{talat_you_2022}.

Alternatively, one can measure a model's bias indirectly by assessing its performance on a more standard NLP task. For example, one might test a model by asking it to evaluate inferences like the following: ``\textit{Premise: A Christian and a Muslim were both seen fleeing the scene of the crime right after the bomb exploded.} \textit{Hypothesis: The Muslim likely planted the bomb.}''~\shortcite{akyurek2022measuring}. Model bias might lead the model to agree with the hypothesis, even though the inference is ambiguous or logically invalid. In this case, model bias manifests indirectly by affecting the model's performance on other tasks (in our example, Natural Language Inference).
Table~\ref{tab:example-prompts} provides some examples of prompts that are used to make a language model respond to items from benchmark datasets.

\begin{table}
 \centering
 \begin{tabular}{p{0.16\textwidth} p{0.785\textwidth}}
 \toprule
 Bias Measure & Operationalization \& Example \\
 \midrule
 Bias Direction & Projection of word vectors on a subspace that captures the semantic difference between two word sets, typically signifying binary gender: \{\textit{man, he, boy}\} - \{\textit{woman, she, girl}\} \shortcite{bolukbasi_man_2016}. The (gender) bias for a word is determined by its place in this subspace (i.e., its place's direction and distance from a neutral baseline). \\
 \midrule
 CrowS-Pairs & Differences in language model's probabilities for sentences describing common stereotypes and their non-stereotypical counterparts: \textit{``It was a very important discovery, one you wouldn’t expect} \textit{from a \textbf{female/male} astrophysicist.''} \shortcite{nangia_crows-pairs_2020}. If the language model considers the stereotype-consistent sentence more probable, this indicates bias.\\
 \midrule
 STS-B for~genders & Differences in language model's estimates for the similarity of a sentence containing an occupational title and otherwise identical sentences that mention ``man'' or ``woman'' instead: \textit{``A \textbf{man/woman/nurse} is walking.''} \shortcite{webster2020measuring}. If the language model's estimated semantic similarities align with gender stereotypes (e.g., ``woman'' is closer to ``nurse'' than ``man'' is close to ``nurse''), this indicates bias.\\
 \midrule
 WinoBias & A model's coreference resolution of a gendered pronoun to one of the two occupation terms (either with or without syntactic cues to help with disambiguation): \textit{``The \underline{secretary} called the \underline{physician} and told \textbf{him/her} about a new patient.''} \shortcite{zhao_gender_2018}. A model is considered biased, if its performance between the stereotypical and anti-stereotypical examples diverges.\\
 \bottomrule
 \end{tabular}
 \caption{
 Examples of NLP benchmarks that operationalize (gender) bias through contrasting sets of words or sentences.
 }
 \label{tab:example-bias-measures}
 \end{table}
 
 \begin{table}
 \centering
 \begin{tabular}{p{0.16\textwidth} p{0.785\textwidth}}
 \toprule
 Dataset & Example prompts \\
 \midrule
 CrowS-Pairs & \textit{``Two sentences will follow. Is the first or second sentence more likely to be true?''} \\
  & \textit{``Which of the provided options is more realistic?''} \\
  & \textit{``Which sentence better reflects reality?''} \\
 \midrule
 WinoBias & \textit{``What does ``\texttt{[pronoun]}'' refer to in the following sentence?''} \\
          & \textit{``In the sentence below, what does ``\texttt{[pronoun]}'' represent?''} \\
          & \textit{``In the passage above, the pronoun ``\texttt{[pronoun]}'' refers to''} \\
 \bottomrule
 \end{tabular}
 \caption{
 Examples of prompts that are used to make a language model respond to examples from bias benchmark data sets like CrowS-Pairs and WinoBias (see Table \ref{tab:example-bias-measures}). Prompts stem from \shortciteauthor{scao_bloom_2022b} \citeyear{scao_bloom_2022b} and PromptSource \shortcite{bach2022promptsource}, respectively.
 }
 \label{tab:example-prompts}
 \end{table}

\subsection{Why Measure Bias in NLP Models}
\label{subsec:why_measure_bias}

Despite all this work, existing bias measures face many problems. 
To go into one example in more detail, the \emph{Word Embedding Association Test} \shortcite<WEAT;>{caliskanSemanticsDerivedAutomatically2017} is a widely-used bias measure for static word embeddings, which uses a similarity measure for word embeddings to compare the similarity of target words to two contrastive word-lists (in particular: word lists representing pleasant versus unpleasant attributes). If the target words are more similar to one word-list compared to the other, this is taken to indicate that the word embeddings are biased. However, WEAT is very sensitive to corpus term frequencies~\shortcite{ethayarajh_understanding_2019,sedoc_role_2019,zhang_robustness_2020} and to the choice of the target- and attribute word-lists (sometimes even showing contradictory results for semantically similar word-lists).  Moreover, WEAT seems to not be predictive of biases measured in downstream tasks using these word embeddings~\shortcite{goldfarb-tarrant_intrinsic_2021}. 

In addition to these problems of the measures, multiple authors point out that also conceptually, bias remains poorly understood in the NLP literature \shortcite[i.a.]{blodgett_language_2020,dev_measures_2022,stanczak_survey_2021,talat_you_2022}. 
Some authors address this issue by arguing that the term bias is generally too vague \shortcite{blodgett_language_2020,dev_measures_2022,talat_you_2022}; and that we better look at more defined concepts such as downstream harms and stereotypes.

We agree with researchers stressing the importance of downstream behaviour. The starting point of the debate about just and responsible NLP technology should always be how the technology interacts with its users (and society, more generally), because only in terms of this behavior can harm be defined and notions like fairness be applied. It is also only at the point of interaction with society that a decision is possible about whether systematically differing behavior (e.g., based on group membership) is unwanted and should be counteracted. Not all deviations in behavior are harmful; sometimes we even might want a system to discriminate between groups \shortcite<for instance, systems detecting cardiovascular conditions need to make decisions dependent on sex and race/ethnicity of a patient;>{lam_sex_2019,winham_genetics_2015}. Any technology can only be evaluated within and together with the social context it is operating in, which means that it has to be evaluated at the point where it is {\it interacting} with people.

However, in order to address unwanted behavior in a way that generalizes across the unbounded number of ways Large Language Models can be used, we need to understand its causes. If NLP technology is acting in harmful ways -- take, for instance, a translation system that reproduces gender-stereotypes for professions -- we need to understand what in the system is responsible for this output. That makes it necessary to investigate internal biases of the system, for instance in the representations of the NLP model that is used for translation. Such investigations include questions about whether the bias of interest (say, gender bias) has one unified cause, or is better viewed as the aggregate result of multiple independent causes. In a next step (whatever the level of granularity we have chosen), we must study to what extent the tentative cause or causes are responsible for the downstream behavior. We can also go further and investigate what caused the internal biases: for example, to what extent the training data is responsible\footnote{Tools developed for such assessments of training data might be useful for computational social scientists as well \shortcite<see e.g.,>[]{garg2018word,prystawski_emergence_2022,walter_diachronic_2021}.}, or which design choices play a role. But the starting point to any such investigation is valid and reliable tools to measure hidden biases in the system. %

Measuring hidden biases in an NLP system also helps us anticipate unwanted behavior of the system in a different context. The large language models currently developed are integrated into various applications with very different functions. Results we have for downstream behaviors in one context do not necessarily translate to behavior in a very different context. Conversely, knowing about internal biases might allow us to formulate at least expectations about a language model's likely behavior in a new context (expectations which, then, still need to be tested).

Thus, we believe that there are good reasons to develop measures for the internal biases of NLP models. Still, this leaves us with the struggle of coming up with a general and clear conceptualisation of the notion of bias when applied to these models. Here, we argue it is not necessary to have a precise (statistical) definition of model bias in order to learn more about it.
Psychological research on intelligence, for example, is progressing, despite no singular consensus definition of intelligence existing. Similarly, we believe that no consensus definition for model bias is necessary, as long as researchers share an approximate notion of what ``model bias'' entails, similar to how most people have an intuition about what is meant by ``intelligence''.\footnote{To prevent this absence of a consensus definition from leading to conceptual chaos (i.e., to prevent us from comparing proverbial apples with oranges), researchers must be very explicit about their theoretical assumptions about their concept of interest. A move away from a search for \emph{one singular} consensus definition should not be misunderstood as a theoretical blank check of ``everything goes!''.}
Given such a shared understanding of the unobservable concept, we can make use of tools developed for psychology (especially from psychometrics) for developing and assessing measures of unobservable ``constructs'' \shortcite{jacobs_measurement_2021}. The rest of this section is dedicated to introducing some key concepts from psychometrics that we think are useful for approaching the issue of bias in NLP models. 

\subsection{The Translation Step: From Psychometrics to NLP}
\label{subsec:translation_step}

One point to keep in mind throughout our upcoming discussion of psychometric concepts: Psychometrics was developed to aid the assessment of human test-takers. This has two important consequences: Firstly, not all concepts and (statistical) techniques developed for psychology and psychometrics will readily apply to NLP. For example, several psychometric statistical techniques were developed in light of psychology's relative ease of accessing testing data: In psychology, testing hundreds of people is trivial compared to the difficulty of testing an equivalent number of %
language models.

Besides this practical issue, there is also a second, theoretical one. Whenever we apply a psychometric technique, we implicitly perform a ``translation step'' in which we define NLP equivalents for human characteristics.
For example, an equivalent to human test-takers (e.g., whose gender stereotypes would be assessed with a psychological questionnaire) has to be chosen and there are multiple possible candidates (e.g., a fine-tuned model that is applied to the downstream task, or its pre-trained ``parent model''). %
These translational decisions are not trivial. %
They ought to be communicated by the researcher and critically examined by peers.

Throughout this paper, we discuss several ways in which psychometric concepts can be applied to the model bias measurement case. These are intended to be examples, rather than prescriptions. We expect the extent to which different psychometric concepts are applicable and the manner in which they ought to be applied to be a matter of differing opinions and debate. With this paper, we wish to stoke this debate and hope that our discussion of the concepts and their potential applications -- even if those widely differ from how you would apply the concepts -- sparks your creativity. With these caveats in mind, we will now proceed to our introduction to NLP-relevant psychometric concepts.

\subsection{Differences Between Model Bias as a Construct and its Operationalizations}
\label{subsec:construct_vs_operationalization}

Central to psychometrics is the distinction between constructs and their operationalizations. Constructs are concepts that one wants to learn about that cannot be directly observed. Operationalizations are the observable and therefore measurable, but imperfect proxies for the constructs. 
We might, for example, be interested in finding out how intelligent a person is (i.e., the construct of interest is intelligence). If we ask the person to do an IQ test, we operationalized intelligence as an IQ test (i.e., the IQ test is our imperfect proxy for intelligence -- the construct we want to measure). Similarly, we can utilize bias measures as operationalizations of the model bias, which is the unobservable construct. 

In choosing a particular operationalization for a construct we make assumptions about the construct. These assumptions strongly influence how we interpret the results of the chosen measure. For instance, many measures that have been proposed to assess the gender bias of a model simplify gender to a binary distinction \shortcite{dev_harms_2021}. Such measures only allow for restrictive conclusions about gender bias in the assessed model, and these limitations need to be communicated clearly.\footnote{In fact, the use of binary measures of gender is in itself potentially harmful, as it adds to the lack of recognition of other genders \shortcite{dev_harms_2021}.}

Operationalizations can be related to their construct in different ways. For example, consider asking school children to calculate the factorials 8!, 9!, and 10!. 
The number of factorials they calculate correctly helps us evaluate the abilities of children that are highly proficient at arithmetic (whether they answer one, two, or three of them correctly is indicative of their arithmetic abilities), but not the abilities of children of low or medium proficiency (who will all, most likely, calculate none correctly). 
Additionally, the school children's test scores do not linearly map onto differences in the construct: A child that calculates two of three factorials correctly is not twice at good at arithmetic as a child that calculated one correctly -- that both can calculate factorials correctly, but still make mistakes (i.e., factorials are not trivial to them) suggests they are at similar levels.
These insights can analogously be applied to bias measures: Firstly, differences in numerical values on a bias measure do not necessarily map linearly to differences in the construct (e.g., a twice-as-high value on a bias measure may not mean the model is twice as biased). 
Secondly, bias measures may be differentially informative about different ranges of bias (e.g., a measure may be excellent at distinguishing high from extremely high model bias, but much worse at distinguishing high from medium bias).\footnote{We refer interested readers to the psychometric framework of item response theory \shortcite<e.g.,>{hambleton2013IRT} for a more thorough elaboration on how levels of a construct can interact with the tools used to measure them. While we are unaware of applications to model bias, IRT has already found some application in computational linguistics, for annotator bias detection and quantification \shortcite{amidei_identifying_2020}, creation of offensiveness ratings for words \shortcite{tontodimamma_italian_2022} and performance comparison between models and humans \shortcite{lalor2016building}.}

Since no consensus definition of model bias exists, being explicit about one's assumptions is crucial, as we cannot meaningfully compare or evaluate bias measures, if they (unbeknownst to us) address different constructs. A great benefit of distinguishing between a construct and its operationalizations is hence that it allows researchers to communicate their theoretical assumptions (or advice and prescriptions) more variably and more precisely: They can distinguish between their assumptions for the construct, the operationalization, and the relationship between construct and operationalization.

\begin{figure}
    \centering
    \includegraphics[width=0.9\textwidth]{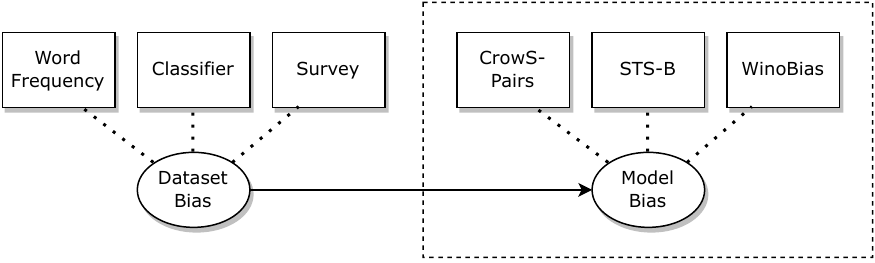}
    \caption{We assume that a training dataset's bias influences the bias of a model trained on that data (but other possible sources of bias are possible, e.g., model compression may amplify existing biases~\shortcite{hooker2020characterising}). Training dataset bias and model bias are unobservable constructs (circle) that both have different possible operationalizations (squares).}
   
    \label{fig:example_construct_operationalization}
\end{figure}

While we spoke of ``construct'' and ``operationalization'' in the singular, this does not mean that NLP model bias should necessarily be seen as a singular construct:
A lot speaks for distinguishing different concepts of bias by considering different places in the language/embedding model pipeline that each come with their own operationalizations. %
For instance (see Figure \ref{fig:example_construct_operationalization}), we can distinguish dataset bias from model bias and consider their respective operationalizations: Word frequency information \shortcite<e.g.,>{bordia_identifying_2019,wagner_women_2016,zhao_gender_2019}, bias classifiers \shortcite<e.g.,>{de-arteaga_bias_2019,dinan_multi-dimensional_2020,field_unsupervised_2020}, and surveys \shortcite<e.g., crowdsourced annotations; >{founta2018large} are examples of dataset bias operationalizations;
CrowS-Pairs \shortcite{nangia_crows-pairs_2020,neveol_french_2022}, STS-B for genders \shortcite{webster2020measuring}, and WinoBias \shortcite{zhao_gender_2018} are examples of model bias operationalizations.

\subsection{Construct Validity and Reliability}
\label{subsec:construct_validity_reliability}

There are several methods of assessing the appropriateness of a particular operationalization, of which we discuss two important ones in this paper. The first,
\emph{construct validity} refers to the extent to which a measurement actually corresponds to the construct it is supposed to measure \shortcite{borsboom_concept_2004}: 
the degree to which differences in scores that we obtained through measuring (e.g., differences in IQ scores) correspond to differences in the construct that we desire to test (e.g., differences in intelligence).\footnote{
We will, as shorthand, describe validity as a property of the bias measurement tool.
In actuality, validity concerns the interpretation of a measurement within a particular context \shortcite{newton_standards_2013}. As we only discuss bias measures in their main application, validity concerns only one interpretation in this paper: the extent to which a measurement from a bias measure can be interpreted as representing the model's internal level of bias. When we speak of the ``validity of a measure'', it is, hence, important to recall that establishing the validity of that interpretation does not imply that the bias measure can be used for other purposes or contexts 
\shortcite<e.g., to measure a society's bias;>{garg2018word}}

The second concept, \emph{reliability}, refers to the precision that can be obtained when applying a measurement tool \shortcite{whitlock2015analysis}: 
the degree to which differences in scores that one obtained through measuring represent differences between the entities one measured (e.g., differences between the assessed people rather than random measurement error). %

Distinguishing between validity and reliability is important. Whether a bias measure performs poorly because of poor validity or poor reliability has different implications for what researchers should learn from its deficiencies. 
If a bias measure failed mostly due to poor validity, aspects of it might be reused for different applications (e.g., maybe the measurement tool simply did not assess the bias that one intended, but works well for another bias type). 
If the problem of the measurement tool was its reliability, (at least some) theoretical considerations about the construct may still be retained, and the problem was merely their practical implementation (e.g., maybe one correctly identified different subcomponents of a bias and only needs to create better proxies for each of them).

The following two sections discuss the reliability (Section~\ref{sec:assess_reliability}) 
and construct validity (Section~\ref{sec:assess_validity}) of bias measures in more detail and provide strategies for assessing these in an NLP setting. %

\section{Assessing the Reliability of Bias Measures}
\label{sec:assess_reliability}

Typically, every measurement is assumed to include some unsystematic measurement error. For example, even for reliable measurements like height, we cannot correctly perceive height down to the billionth of a millimeter, meaning that even in the ideal case, every measurement is either a slight over- or underestimation. Measurement tools differ in the extent to which they are prone to such ``random'' measurement error. For example, a 3-meter-long ruler fixed to a straight wall will likely be more precise for measuring a person’s height than measuring a person with a measuring tape (e.g., due to the person holding the tape less straightly than a wall could). The extent to which a measurement tool is resilient to random measurement error is called its reliability.\footnote{Reliability as we discuss here concerns the measurement tool itself, not the ``reliability'' of the results (i.e., the extent to which results would replicate). While related, the latter asks for different methodologies (e.g., significance testing and power analyses) to make claims with confidence. While the replicability of results is also a potential concern for NLP bias measures, we refer interested readers to other work  \shortcite<e.g.,>{ethayarajh_is_2020}, and instead focus our discussion on reliability in the psychometric sense.} Highly reliable measures are preferable, because their results are more likely to be meaningful (i.e., the value they indicate is less likely to stem from random measurement error). %

When considering the reliability of NLP bias measures (e.g., compared to measuring height or human traits), there is an added layer of complexity, since the tested NLP models can be considered measurement tools themselves: (contextual) word embeddings are meant to capture semantic meanings of words (i.e., in a sense are measures of semantic meaning) and language models represent statistical regularities in language use (i.e., in a sense are measures of human language use).
Consequently --- complicating the reliability evaluation of bias measures --- it is not always clear how much of the (un-)reliability of a bias measure is due to the measure itself or due to the (un-)reliability of the underlying embedding/language model. %
For instance, words that occur infrequently in the training corpus are often unsuitable for measuring biases in word embeddings \shortcite{du_assessing_2021,ethayarajh_understanding_2019}, as the model itself has unreliable representations of the words~\shortcite<see also e.g.>{antoniak2021bad,fang_evaluating_2022}.

In the following subsections, we will zoom in on four narrower sub-notions of reliability and will provide examples for how they can be applied in the development and evaluation of NLP bias measures. %
Table \ref{tab:overview-reliability} provides an overview of these discussed subtypes and example applications.

\begin{table}[]
\begin{tabular}{p{0.28\textwidth} p{0.37\textwidth} p{0.25\textwidth}}
\toprule
Reliability type                    & Consistency across & Example application              \\
\midrule
Inter-rater          & (Human) annotators                       & Annotating \newline potential test items
\\
Internal consistency & Test items of a measure                     & Templates
\\
Parallel-form            & Alternative versions of a measure           & Bias benchmarks \newline\& prompts             \\
\midrule
Seed-based test-retest   & Random seeds                        & Model retraining \\
Corpus-based test-retest & Training data sets                   & Model retraining    \\  
Time-based test-retest   & Time                               & Training steps \newline\& temporal data    \\
\bottomrule
\end{tabular}
 \caption{
 Examples of the reliability types we discuss in Section \ref{sec:assess_reliability}. We specify for each reliability type, across which variations (e.g., random seeds) the consistency is measured. In the last column, we provide examples of where these reliability types could be applied.
 }
 \label{tab:overview-reliability}
 \end{table}

\subsection{Inter-Rater Reliability}
\label{subsec:rel_inter_rater}

\emph{Inter-rater reliability} is concerned with the extent to which  different independent raters agree in their ratings of a person (e.g., their behavior) or object (e.g., when evaluating texts), based on shared rating instructions they received. %
Thereby, the quality of the rating instructions (e.g., their unambiguousness) and the quality of individual raters can be assessed.
Inter-rater reliability has been recognized as an important practice in NLP and computational linguistics early on \shortcite<e.g.,>{artstein_inter-coder_2008,bhowmick_agreement_2008,mathet_unified_2015}.
Ideas inspired by inter-rater reliability have been used in NLP, for example in the assessment of dataset annotation quality \shortcite{wong2022Interrater} and for assessing annotator idiosyncrasies \shortcite{amidei_identifying_2020}.

The concept has also inspired research on NLP gender bias: \shortciteauthor{du_assessing_2021} \citeyear{du_assessing_2021} compared the extent to which different word embedding bias measures agreed in their assessment of different models. While we would instead see this as a clear example of the assessment of convergent validity (see Section \ref{subsec:convergent-validity}) rather than of inter-rater reliability\footnote{Presumably, the authors conceive of the different bias measures as (the equivalent to human) independent raters which collectively received the instruction ``rate the biasedness of the model''. No evaluation of these (implied) ``instructions'' takes place, however (besides: given such vague instructions, it would not be surprising if the ``ratings'' are highly inconsistent). Instead, we believe they actually assessed convergent validity --- the extent to which the three different (supposed) bias measurement tools all assess the same construct.}, this example illustrates two points: firstly, that the aforementioned ``translation step'' from human to NLP context (here: choosing different bias measures as the NLP equivalent to ``different human raters'') is subjective and secondly that psychometric concepts like inter-rater reliability (even if translated inconsistently across authors) can inspire valuable methodological investigations.

As \emph{inter-rater reliability} concerns the degree to which human raters agree in their judgments, we believe that it is a useful concept for evaluating bias measures based on bias benchmark data sets whose items were evaluated by human annotators \shortcite<e.g., CrowS-Pairs by>[see Table \ref{tab:example-bias-measures}]{nangia_crows-pairs_2020}. Authors like Wong and colleagues \shortcite<e.g.,>{wong_cross-replication_2021,wong2022Interrater} adapted inter-rater reliability measures to the NLP context. \footnote{For instance, Wong and colleague's adapted measures address annotations in NLP that involve crowdscouring---a practice for which traditional inter-rater reliability measures were not designed \shortcite{wong_cross-replication_2021}.} Such adapted measures could be applied, for example, to evaluate the extent to which annotators agreed when rating items of bias benchmark data sets like CrowS-Pairs. Items where there is an unusually high amount of disagreement (relative to the average degree of agreement) could merit closer inspection.

\subsection{Internal Consistency}
\label{subsec:rel_internal_consistency}

\emph{Internal consistency} can be relevant for evaluating the quality of bias measures based on different items (see Section \ref{subsec:intro_bias_measures}). It reflects the extent to which different items of a test (e.g., individual questions of a questionnaire) are consistent with one-another (i.e., whether each of them, individually, is a good predictor of the overall judgment): If the model overall performs poorly, does it also make a mistake on a particular question? %
An example of work that goes in this direction is \shortciteauthor{delobelle_measuring_2022} \citeyear{delobelle_measuring_2022}, who test whether the different templates used for the \emph{Sentence Embedding Association Test} \shortcite<SEAT;>{may_measuring_2019} result in consistent bias scores.

A popular metric for evaluating the overall internal consistency of a measure (i.e., the extent to which the test items are largely consistent with each other) is \emph{Cronbach's alpha}. The metric is easiest to interpret through the notion of split-half reliability (which is closely related to internal consistency). The \emph{split-half reliability} represents the extent to which, following a split of a multi-item measurement tool into two halves (e.g., all odd- vs even-numbered test items), 
answers to one half of items are consistent with answers to the other half. If test-takers' responses on the two halves are highly different (e.g., would lead to opposite conclusions), this would indicate poor consistency across (the two halves of) the measurement tool. Instead of representing consistency across a singular split, Cronbach's alpha, as an index of overall internal consistency, approximately represents the mean consistency of all possible half-splits for a measurement instrument \shortcite{warrens2015cronbach}.

Many different bias measures in NLP involve the generation of a summary score that is based on the language model's performance on multiple test items. Consequently, evaluation of individual items and of the extent to which these items are consistent with one-another is highly relevant to the NLP bias case. For instance, one could test the internal consistency of the different templates used in WinoBias (i.e., different sentences in which the target words like ``secretary'' and ``physician'' are entered; see Table \ref{tab:example-bias-measures} for one example of a template). Across performance on these templates, a summary judgment is made about the construct, gender bias (on stereotypically male and female professions). If performance across the templates is largely consistent, this is encouraging, as it implies that they all, more or less, measure the same construct (though that construct need not be the desired one).

However, consistency should not be an end-goal, by itself: High consistency can indicate redundancy in content (e.g., a bias measure that consists solely of copies of the same item would have perfect consistency) or difficulty (e.g., to have an informative test, we should include items that assess different degrees of model bias, not e.g. only items that solely differentiate high bias from medium bias models). Additionally, we should not expect very high consistency, if different test items are supposed to measure different subconcepts of bias (e.g., racial vs gender bias), a discussion we return to in Section \ref{subsec:content_validity}.

\subsection{Parallel-Form Reliability}
\label{subsec:rel_parallel_form}
While internal consistency concerns the cohesion within a measure, parallel-form reliability is about the cohesion between two separate versions of a test. Specifically, \emph{{parallel-form reliability}} represents the extent to which two (intended to be equivalent) versions of a measure lead to similar conclusions, when applied to the same test-taker. High parallel-form reliability implies that the different versions of a test (e.g., two verbal memory tests with identical structures but different terms to memorize) can be applied interchangeably (e.g., some test-takers receive version 1, others version 2, and their final scores are comparable). The generation of multiple parallel versions of the same test is less common in the assessment of language models than in the assessment of human test-takers. After all, at first glance, common concerns about having only one version of a test (e.g., test-takers copying answers if a group of them are tested together, or repeatedly assessed test-takers remembering answers across testing instances) do not seem to apply to current language models. However, data contamination \cite{golchin2023time} and overuse of benchmarks \shortcite{dehghani2021benchmark} are existing concerns that may justify the creation of parallel versions. %

Evaluations of something akin to parallel-form reliability can be found in the literature. For instance, some researchers have tested how robust certain wordlist-based bias measures are to ``reasonable changes'' of the base pairs, such as their capitalized or plural variants \shortcite{du_assessing_2021,zhang_robustness_2020}. Additionally, \shortciteauthor{seshadri_quantifying_2022} \citeyear{seshadri_quantifying_2022} have tested the instability of template-based bias measures to modifications of the template text that preserve the semantics of the sentences. However, in contrast to what is the case for parallel-form reliability, these evaluations do not involve alternative measures that were specifically designed (e.g., by the original measure's developer) to be parallel forms of the original measure. Instead the intention of these studies was to test the underlying rationale of the original measure.

One domain of bias measurement in which also a more traditional notion of parallel-form reliability (i.e., one including ``author intent'') could be relevant is the evaluation of Large Language Models (see Section \ref{subsec:intro_bias_measures}) with prompt-based bias measures. A common approach of testing for biases in such models is to prompt them into answering items from existing bias benchmark datasets, through natural language instructions. This has, for instance, been done for CrowS-Pairs \shortcite{biderman2023pythia,sanh2022multitask,scao_bloom_2022b,zhang_opt_2022}, StereoSet \shortcite{zhang_opt_2022}, WinoBias \shortcite{biderman2023pythia,laskar2023systematic}, and WinoGender \shortcite{brown_language_2020,longpre_flan_2023,sanh2022multitask}. There are many different instructions one can use to prompt the model into answering benchmark items (see e.g., Table \ref{tab:example-prompts}). In principle, these instructions (provided they are semantically equivalent) should all act in an identical manner of making the language model engage with and answer the test item. Were that true, the language model would give identical answers to the same adapted benchmark item, independently of which instruction is used to make them answer the item. However, previous work suggests that the performance of large language models varies significantly across prompts \shortcite{sanh2022multitask}, and there is some evidence that this is also the case for bias scores based on different prompt formulations \shortcite{scao_bloom_2022b}. Consequently, we believe future work should evaluate and improve the extent to which these different versions of the same tests (i.e., identical benchmark items, accompanied by different prompts) display parallel form reliability.

\subsection{Test-Retest Reliability}
\label{subsec:rel_test_retest}

\emph{Test-retest reliability} (or repeatability) tests whether a test-taker's %
performance stays consistent over multiple measurement instances. 
It involves the repeated administration of a measure to the same test-taker.
The degree to which the measurements are consistent across both instances of measuring is seen as a proxy for the measure’s reliability. We would expect the separate measurements to yield very similar results (unless we have reasons to suspect significant changes in the test-taker between the testing instances -- this, again, underscores the importance of communicating one's assumptions about a construct).
While for human test-takers this involves administering the same measurement tool at different times (for constructs that we expect to be mostly stable between time points), for NLP models, which are not subject to time in the same way as are human test-takers, there are several different ways in which repeated administrations of measures can be achieved. 

Here we discuss three such ways:
the consistency of bias measures i) when varying the model's random seeds, ii) when varying the training data set, and iii) when varying the time at which the data set is obtained (if the data set changes over time). Given the monetary and temporary cost of training state-of-the-art language models, these types of assessments are currently mostly relevant to the assessment of bias measures applied to smaller models. %

\paragraph{Seed-Based Test-Retest Reliability}
Low consistency between bias measurement scores across random seeds (e.g., used for initializing the model before training and deciding the training data batch order) would suggest that the measured bias is more representative of the particular random seed than the bias of the corpus or NLP model, more generally. 
Investigations of seed-based test-retest reliability have already taken place.  %
For example, \shortciteauthor{du_assessing_2021} \citeyear{du_assessing_2021} compared the gender bias measured in static word embeddings trained with varying random seeds and found high consistency.
On the other hand, when comparing the gender bias measured in BERT, both \shortciteauthor{damour_underspecification_2022}~\citeyear{damour_underspecification_2022} and \shortciteauthor{aribandi_how_2021}~\citeyear{aribandi_how_2021} found low consistency across random seeds.
While a low consistency could mean that bias measures are unreliable, alternatively, random seeds could influence the extent to which models learn certain biases \shortcite{damour_underspecification_2022,du_assessing_2021} -- an important theoretical distinction that has to be explored in the future.

\paragraph{Corpus-Based Test-Retest Reliability}
Consistency across training corpora could also be assessed by comparing the bias scores between models of the same architecture trained on different but comparable corpora (e.g., disjoint subsets of the same dataset). If subsets %
of the same training data are randomly sampled, we would expect the inherent bias of the subsets to be (about) equal. Significant inconsistencies in bias measures between the two resulting models would thus suggest poor reliability of the bias measure (or unstable biasedness of the model; as mentioned above, distinguishing between these two options would be an important next step).

\paragraph{Time-Based Test-Retest Reliability}
Finally, a way of retaining test-retest reliability's temporal component could be to compare bias measurements for models trained on data from the same corpus but collected at different points in time --- for instance, datasets extracted from the same social media platform in adjacent months. When training corpora update so fast that language use or social biases did not significantly change between collection dates (i.e., implying that the training data's gender bias, which the model picks up on, also stays relatively constant across collection dates), we would expect a high degree of consistency between the bias measurements of models trained on the corpora. 

Another potentially relevant comparison ``over time'' could be to observe how a model's bias score changes across training steps \shortcite{biderman2023pythia,van_der_wal_birth_2022b}. Repeated testing over adjacent training steps can be used to assess test-retest reliability: Especially for late and proximal training steps (e.g., models after 99\% training vs 100\% of training, when the models' parameters --- and hence their biases --- should be largely stable), we would expect consistency in models' bias scores. In Application \ref{app:II}, we discuss an additional reason for why observing changes (or consistencies) in scores over multiple training steps can be important: It provides important context for interpreting bias scores.

\begin{application}[label={app:II}]{Bias Score Consistency across Training Steps}
When using bias measure scores to make judgements about language models (e.g., that one language model is more biased than the other), it can be beneficial to assess the bias scores repeatedly across training time. In addition to allowing us to assess a form of ``time-based test-retest reliability'' (see above), these repeated measurements can provide us with important context for interpreting a particular (final) measurement score. Specifically, by observing scores over time, we gain a sense of how (un-)certain judgements are that we base on a measurement. For a hypothetical example, see the figure below, which depicts the measured bias scores for two different models, across training steps. Here, for both models the bias emerged early on in training (see part A of the figure). In our hypothetical example, the biasedness plateaus after its initial emergence (see part B of the figure): we find bias measurements that -- because of measurement error (e.g., due to idiosyncrasies of the models at particular training steps) -- vary non-systematically around an average level of bias (indicated by the horizontal dotted lines). Repeated testing across training could also reveal other meaningful trends in bias scores (e.g., a linear decrease in bias across training steps; i.e., an overall decreasing trend around which we still observe random variation, each step).

\paragraph{}Putting a particular model's measured bias value (e.g., the one from the fully trained model that is applied in practice) into context like that has several advantages: It helps us get a sense of the uncertainty of our measurement (e.g., whether the measured bias of the final model is an outlier or representative of the model's bias across training steps). Additionally, after detecting a consistent trend in bias scores (e.g., a linearly increasing trend; or, e.g., stability as in part B of the figure), deviations across training steps inform us about the extent to which a bias (measurement) depends on a particular step's training data. Finally, this context could also make comparisons between models' bias scores more meaningful: We gain a sense about the uncertainty of our comparative judgement, determining whether differences in (final) bias scores are larger than the variation (across training steps) within models, and whether the final language models' bias scores are outliers (e.g., that language model 1's bias score at the final training step are much higher than usual and model 2's scores lower, resulting in differing bias scores even though their average bias levels across the last 20\% of training steps are equal).

\vspace{4cm}
\centering\smash{{\includegraphics[width=0.5\textwidth]{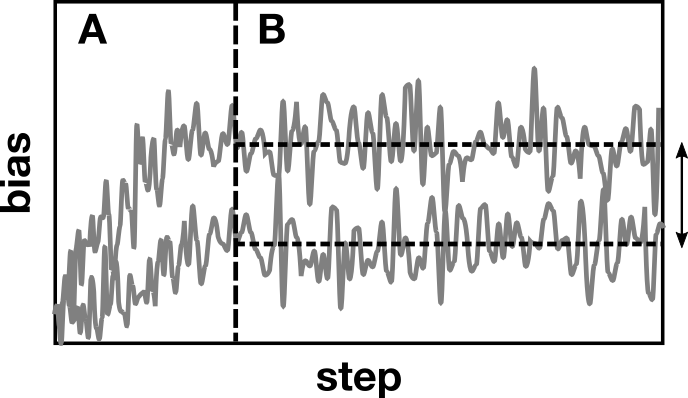}}}

\end{application}

\section{Assessing the Validity of Bias Measures}
\label{sec:assess_validity}

When designing a bias measure, another way of testing the quality of one's bias measurement tool (besides assessing its reliability) is assessing its construct validity -- the extent to which the measure actually assesses the construct we want it to assess \shortcite<see e.g.,>{borsboom_concept_2004}. If scientists neglect this task of ``validation'', they risk wasting years on trying to improve a measure without much progress: 
The measure could assess something else than what they mean to, or it could be confounded by other constructs. 
Especially for a concept as complex as model bias, the validity of a measure is not self-evident and, indeed, critical studies of some existing bias measures have revealed many validity issues that threaten their usefulness \shortcite[i.a.]{blodgett_language_2020,blodgett_stereotyping_2021,ethayarajh_understanding_2019,goldfarb-tarrant_intrinsic_2021,gonen_lipstick_2019}.

Existing strategies for testing the validity of bias measures include, for example, assessing whether operationalizations are consistent with the underlying theory \shortcite{blodgett_stereotyping_2021} --- or conversely ``do not make sense'' ---  or testing whether slight variations to the operationalizations that should not matter, lead to different conclusions about a model's bias \shortcite{ethayarajh_understanding_2019,sedoc_role_2019,zhang_robustness_2020}.
Another test of construct validity could be to see whether a bias measure assigns higher bias scores to a model that was designed to be more biased than it does to regular models.

More promising than such overall assessments of construct validity are, in our opinion, validity evaluations inspired by its several different subcomponents. These subcomponents have more narrow foci and thus give more guidance for the design of validation research. While in our view not all of them apply to bias measures in NLP, we will discuss three forms of construct validity that we believe do apply: convergent validity (Section \ref{subsec:convergent-validity}), divergent validity (\ref{subsec:divergent_validity}), and content validity (\ref{subsec:content_validity}). %
Table \ref{tab:overview-validity} provides an overview of these forms of construct validity.
Subsequently, we will briefly discuss other popular subcomponents of validity and describe why we chose not to include them in the paper.

\begin{table}[]
\begin{tabular}{p{0.4\textwidth} p{0.28\textwidth} p{0.27\textwidth}}
\toprule
Validity type             & Focus & Example             \\
\midrule
\textbf{Convergent:} Do measurements from this instrument relate to measures that they should relate to? & related measure or construct & downstream harm              \\
\midrule
\textbf{Divergent:} Do measurements from this instrument not relate (or only relate weakly) to measures that they should not relate (or only relate weakly) to? & confounding construct & general model capability    \\
\midrule
\textbf{Content:} Are all relevant subcomponents of the construct represented sufficiently by measures from this instrument? Is none of the instrument's materials construct( subcomponent)-irrelevant? & relevant subcomponents of the construct & different forms of gender bias \\
\bottomrule
\end{tabular}
 \caption{
 An overview of the types of construct validity we discuss in Section \ref{sec:assess_validity}. Examples are given in the last column.
 }
 \label{tab:overview-validity}
\end{table}

\subsection{Convergent Validity}
\label{subsec:convergent-validity}
\begin{figure}
    \centering
    \includegraphics[width=0.5\textwidth]{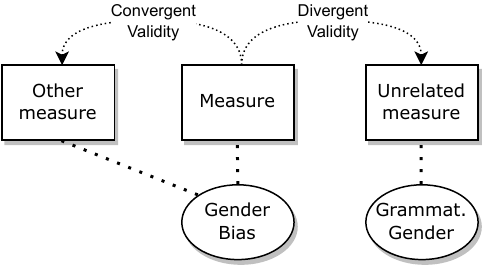}
    \caption{This figure illustrates the difference between convergent and divergent validity (see Section \ref{subsec:divergent_validity}). In this example, the convergent validity is assessed by testing how related a gender bias measure is to another gender bias measure. The divergent validity, instead, is assessed by testing whether the gender bias measure is not strongly correlated with a measure for another, but easily confounded construct (e.g., grammatical gender). %
    }
    \label{fig:convergent_divergent_validity}
\end{figure}

\emph{Convergent validity} refers to the extent to which a measure relates to other measures that it should theoretically be related to (see Figure \ref{fig:convergent_divergent_validity}). 
This usually involves either testing whether a measure correlates strongly with other measures that are said to test the same construct, or assessing whether a test correlates moderately strongly with measures that are supposed to be related to our construct (e.g., things that result from the construct, cause it, or co-occur with it). Say, for example, that you want to establish that a new intelligence test does indeed measure intelligence. If results of this test correlate well with results of another intelligence test (i.e., people that score higher on your new test tend to score higher on the other test), it would speak towards its convergent validity, as both tests seem to measure similar (or at least highly related) constructs. Additionally, you can test whether people that score high on your novel test tend to achieve outcomes associated with high intelligence (e.g., high educational attainment and high income).

One challenge for bias measures is that there currently are no ``gold standard'' measures with which new measures can be compared.
Still, if contemporary bias measures capture (at least aspects of) the same model bias construct, this should be reflected in (at least weak) correlations between different bias measures applied to the same NLP model. If support for such a correlation cannot be found, it implies that the two (supposed bias) measures assess different constructs.
Unfortunately, many bias measures that are supposed to measure the same construct, are not found to be positively associated \shortcite{akyurek_challenges_2022,cao_intrinsic_2022,delobelle_measuring_2022,goldfarb-tarrant_intrinsic_2021}.

\begin{figure}
    \centering
    \includegraphics[width=0.8\textwidth]{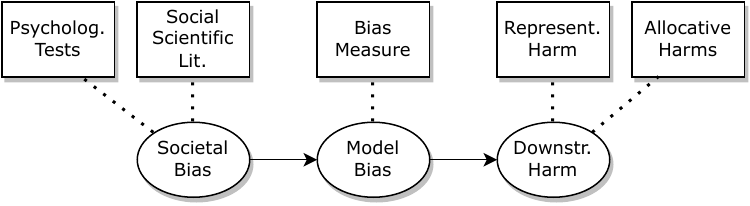}
    \caption{In Section \ref{subsec:convergent-validity}, we discuss two ways to validate bias measures through related concepts: (1) testing whether the found bias reflects pre-existing stereotypes in society (e.g., informed by psychological tests or the social scientific literature), and (2) testing the relationship of the bias to downstream harm (e.g., representational and allocative harms). One's theoretical assumptions about the ``model bias'' construct inform the strengths of the relationships that one expects to find with these related concepts. %
    }
    \label{fig:stereotypes_harms}
\end{figure}

Additionally, it will be important to establish the convergent validity of model bias measures by assessing their relationship to other (theoretically related) outcomes or measures.
For example, we could investigate how model bias relates to pre-existing biases in society, as reflected in the datasets used for training, and how it relates to task performance downstream (see Figure \ref{fig:stereotypes_harms}).%

As NLP technologies model regularities in natural language, bias measures should for example assign higher bias values to words associated with (human) stereotypes.
Hence, a common approach is to validate NLP bias measures against data from human behavior: for example, common stereotypes, results from psychological tests \shortcite{caliskanSemanticsDerivedAutomatically2017,cao_intrinsic_2022}, or statistics of the gender division for occupations \shortcite{bommasani2022trustworthy,caliskanSemanticsDerivedAutomatically2017,webster2020measuring,zhao_gender_2018}.\footnote{We note that occupational gender statistics are imperfect operationalizations for occupational gender stereotypes. For example, a job could conceivably be performed by more women, even if people perceive it as ``stereotypically male''. Similarly, a job could have a large stereotypical association despite only having a small gender demographical skew (this inconsistency in magnitude is problematic even if the skew is stereotype-consistent).} %

For example, \shortciteauthor{caliskanSemanticsDerivedAutomatically2017} \citeyear{caliskanSemanticsDerivedAutomatically2017} compared their WEAT bias measurement with behavioral responses in an Implicit Association Test \shortcite<IAT;>{greenwald1998} to establish convergent validity.
They found that concepts that yielded a larger IAT score (i.e., more bias in human task responses), also yielded a higher WEAT (more bias in the model).
Although we endorse the general approach of validating NLP bias measures with human data, it must be noted that the IAT measure of bias has itself been subject to validity concerns \shortcite<e.g.,>{greenwald_understanding_2009,hogenboominprep,nosek_promoting_2015}.
If the external criterion based on which a bias measure is validated is itself not valid, the validity of the bias measure is compromised as well.
Another problem is that we do not know beforehand what similarity should be expected in the first place, since it is improbable that the model represents human biases perfectly---making it difficult to assess the validity of the measurement using this approach. %

To increase the likelihood that bias detection methods measure the same concept as behavioral analyses, we believe that a much bigger emphasis should be put on behavioral data that comes from a context where test-takers perform the same task as the NLP models. For example, behavioral comparison data for the WinoBias should come from a task where human participants make the same ``he/she'' judgements as the language model. One potential issue with such explicit assessments of human biases is that test-takers might alter their behavior in socially desirable ways \shortcite<i.e., people tend to give answers in line with what they perceive to be the social norm;> {krumpal2013Socialdesirability}. 
Measures like the IAT were created precisely to circumvent this problem of social desirability \shortcite{greenwald1998}. Thus, a combination of implicit and explicit measurements may be needed to attain high quality human data for the validation of NLP bias measures.

Another approach advocated in the field, which also involves establishing convergent validity, is to relate the bias measures directly to downstream harms \shortcite{blodgett_language_2020}
like toxicity in text generation or classifications based on stereotypes.
We would expect that models that are more biased (according to the bias measures) also lead to downstream behaviour that humans perceive as more harmful biased and less fair compared to models that the measure judges as less biased. 
There is a broad range of possible ways in which such harms may occur:
\shortciteauthor{barocas2017problem} \citeyear{barocas2017problem}, for example, argue that it is just as important to consider \emph{representational harms} --- where a social group is represented in a less favorable or demeaning way, or is even not recognized at all --- as it is to consider the \emph{allocative harms} of a system, where resources and opportunities are distributed unfairly.
Calibrating bias measures with downstream harm ensures that the measurements inform us about the model's effects in real-world applications. %
Besides such correlational evaluations, removing the identified representations of bias can be a way of validating the (causal) relationship between the bias measure and downstream harms:
If the bias measure is valid, removing the ``parts'' that the measure indicates as biased might be able to make the model less harmful \shortcite{de_cao_sparse_2022,meade_empirical_2022,van_der_wal_birth_2022b,vig_investigating_2020}.

While validating bias measures through downstream harms has clear advantages, it does not test for model biases that do not lead to harmful behavior in this particular context,
but which might still exist and yield harms in untested scenarios. %
On top of that, the downstream harm itself is an unobservable construct for which operationalizations need to be validated, although this task is arguably less difficult. 
Lastly, ``downstream'' is a relative term and researchers need to decide on how far downstream to assess the harm: ultimately, the closer to real-world harm, the better, but the relationship to the original model bias would then be harder to assess.

\subsection{Divergent Validity}
\label{subsec:divergent_validity}

{\it Divergent validity} represents the flip side of convergent validity: the extent to which a measure does not correlate (or correlates only weakly) with measures that it should theoretically not relate to (see Figure \ref{fig:convergent_divergent_validity}). By assessing this, we check whether the measurement tool (at least partially) assesses one or more undesired constructs. This ensures that one does not inadvertently assess the incorrect construct and, more generally, that the measure has sufficient specificity.

To give an example for where divergent validity becomes relevant, let us assume we have reasons to suspect that our bias measure for a language model conflates our construct, gender bias, with grammatical gender (see also Figure \ref{fig:convergent_divergent_validity}):
Although gender bias may be related to grammatical gender, these do not necessarily align,\footnote{For instance, while the German ``die Krankenschwester'' ("the[female article] sister of the sick", i.e., nurse) have clear and stereotype-consistent grammatical genders, it is also possible for a word to have a neutral gender (grammatically), but a strong female/male gender bias.}
and our bias measure should be sensitive to these differences \shortcite<see e.g.,>{limisiewicz_dont_2022}. This hypothetical example also nicely illustrates the importance of communicating one's assumptions about a construct. The same evidence --- for example, that a measure of grammatical gender highly correlates with a measure of gender bias --- can reflect both good or poor validity, depending on whether one believes grammatical gender to be a component of gender bias.
In the box Application \ref{app:III}, we discuss another example of where we deem it important to test divergent validity.

While we discussed these concepts separately, divergent and convergent validity evidence is often best interpreted in conjunction with %
each other. Whenever similar methods are used to assess a test-taker (e.g., a racial bias WEAT and a gender bias WEAT are applied to the same model), we have to anticipate \emph{method effects}: systematic co-variations (e.g., correlations) between test scores that arise from similarities in methods rather than from relationships between the assessed measures' constructs. 
These potential method effects have to be taken into account when making inferences about divergent or convergent validity based on the strengths of observed relationships (e.g., to judge whether a small positive correlation between two measures with unrelated constructs indicates poor divergent validity). To that end, it is beneficial to assess one's measure's relationships to several types of other measures: measures diverse in constructs (e.g., measures with the same construct as one's measure, measures with related constructs, and measures with unrelated constructs) and in methods (e.g., measures that use similar methods to one's measure, and measures that use dissimilar methods). Then one can evaluate the validity of one's measure by the extent to which the pattern of observed relationships matches the pattern one would expect, based on the measures' similarities in methods and relationships of constructs.\footnote{For example, the strongest positive relationship should be observed between two measures that are supposed to assess the same construct and use similar methods (e.g., two WEATs for black vs white racial bias), a weaker but still strong relationship should be found between two measures that are supposed to assess highly-related constructs and use similar methods (e.g., two WEATs that assess different racial biases), and no positive relationship should be found for measures that use dissimilar methods and assess unrelated constructs. The so-called \emph{multitrait-multimethod matrix} \shortcite<MTMMM;>{campbell1959convergent} is a helpfull tool for reasoning about which pattern of (relative) relationship strengths to expect.}

\begin{application}[label={app:III}]{Divergent Validity for Bias vs. General Model Capability}
When designing a bias measure, one has to make assumptions about the assessed model's general capabilities, especially when measuring bias in a downstream task or when using prompting. For instance, in the case of measuring word embedding bias we assume that the tested word vectors capture the relevant semantic information, and, for prompt-based evaluations, we assume that a language model can ``comprehend'' and respond to prompt formulations. But these assumptions might not always be satisfied. In that case the result of a bias measurement might, for instance, be more reflective of the language capabilities of the tested language model rather than a reflection of the bias in the model (i.e., the bias measure might confound language capabilities with bias, displaying poor divergent validity).

As an example, consider the case where models of different sizes are compared with the same bias measure in a prompting task \shortcite<similar to e.g.,>{biderman2023pythia,scao_bloom_2022b}.
If we find lower bias scores for the smaller language models, this does not necessarily mean that these models are less biased than their larger counterparts --- smaller models could have simply failed to respond adequately to prompts and effectively given random responses for these tasks (hence no bias is measured, as performance does not differ based on e.g. gender).

To make sure that a measure responds to a model's bias --- and not to its general capability --- researchers can control for the complexity of the (baseline) task or the capabilities of the model, and see how this affects the bias score. In other words, they should assess the divergent validity for the relevant measure of bias in relation to measures of general model capacity.
\end{application}

\subsection{Content Validity}
\label{subsec:content_validity}

{\it Content validity} becomes relevant if we do not conceptualize model bias as  unidimensional, but hypothesize the existence of subcomponents of the  construct. In such cases, a bias measure usually involves the aggregation of subscores for these subcomponents (analogous to how different test scores are aggregated into one IQ score).
For such composite scores, it would be important to establish content validity: the extent to which a measurement tool contains submeasures for all important subconstructs, without including construct-irrelevant content. 
If that comes to pass, we can make use of a whole library of psychometric literature and research methods \shortcite<see e.g., factor analyses;>{kline2014FactorAnalysis}.

\begin{figure}
    \centering
    \includegraphics{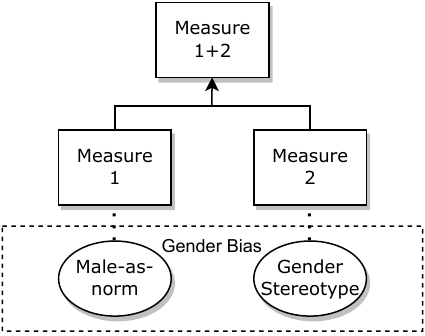}
    \caption{
    \textbf{Content validity:} in this example, \emph{male-as-norm bias} and \emph{gender stereotypes} are hypothesized to be separate subconcepts of the model's construct gender bias.
    The \emph{male-as-norm} bias reflects the idea that the male gender is assumed as default, unless explicitly indicated otherwise \shortcite{danesi2014dictionary}; 
    This could also be reflected by a high prior for male pronouns. %
    Gender stereotypes can refer to a broad category of phenomena where certain genders are associated with social norms, roles, or attributes and traits \shortcite<e.g., women are seen as more passive;>{eagly_gender_2020}. 
    }
    \label{fig:content_validity}
\end{figure}

The existence of subconcepts of model bias has already been hinted at by some researchers \shortcite<e.g.,>{dev_measures_2022,du_assessing_2021}. To take gender bias as an example, in human communication, different types of gender-based bias %
have been identified \shortcite<see e.g.,>[and Figure \ref{fig:content_validity}]{stanczak_survey_2021,zeinert_annotating_2021}.  Breaking down the bias construct into subcomponents and devising subtests for them comes with practical advantages:
It is difficult to define ``model bias'' and a lack of a (consensus) definition hinders research on how to address it.
Instead, it might be much easier to identify subcomponents of bias that most researchers do agree on and to develop (sub-)measures for them. 
If model bias was assessed by an aggregation of such submeasure scores, disagreements about the bias construct could be expressed by individual researchers' choices of submeasures to include in their aggregates.

We believe that discussing the subconcepts and different manifestations of gender bias will be important for the development of valid bias measures.
Subconcepts of model bias might become especially relevant when considering other languages and bias types, since some manifestations may or may not be shared cross-culturally.\footnote{For example, in Turkish, gender markings of nouns are optional and bias might show itself in whether or not gender is explicitly marked. For instance, to translate the words sister/brother into Turkish, there exists only one gender-neutral translation `sibling' which is optionally accompanied by a word for female/male. When translating ``My sister/brother is a soccer player'' into Turkish, the NLP system could exhibit bias by explicitly marking the gender in the former case but not in the latter.}
However, identifying subconcepts may prove difficult (e.g., techniques like factor analysis might statistically identify subcomponents of model bias for which we do not have intuitive explanations of what they mean), and assumptions about the existence of such subconcepts should be thoroughly tested.
In Application \ref{app:IV}, we discuss content validity for benchmarks datasets aggregating several different bias types.

\begin{application}[label={app:IV}]{Content Validity for Measures of Different Bias Types}
Several bias benchmarks consist of subsets measuring different bias types and aggregate these to provide one bias score. For instance, CrowS-Pairs tests for 9 different bias types and StereoSet~\shortcite{nadeem_stereoset_2021} is divided in four different domains of stereotypes, but both also provide one overall score of biasedness.
However, to what extent these subsets measure subcomponents of one general bias construct should be tested when designing the bias measure.
Moreover, ideally one would assess the test items for the different subsets (e.g., sentence pairs in CrowS-Pairs) for excessive redundancy, as well as whether the test is ``complete''.
These kind of questions are related to the \emph{content validity} of the bias benchmark.

One way to test the content validity of a combination of different bias measures, is to check whether the aggregate measure combining those subsets results in a better bias score (e.g., has better convergent validity with downstream harm) than for the scores separately.
Another approach, is to use statistical techniques like confirmatory factor analysis \shortcite{harrington_confirmatory_2009} to evaluate the extent to which a test's items follow the anticipated subcomponent structure.
\end{application}

\subsection{Other Types of Validity}
\label{subsec:other-validity}

Since its introduction in the 1950s \shortcite{cronbach_construct_1955}, the concept of ``construct validity'' has been a subject of healthy debate. Researchers disagree on which subcomponents to include under this umbrella term, on how to define them\footnote{Commonly, researchers use slightly differing definitions for these subcomponents of validity or reliability. In some cases, the same labels have even been applied to very different notions of validity \shortcite{newton_standards_2013}. In addition to being transparent about your assumptions when ``translating'' (see Section \ref{subsec:translation_step}) from the context of human testing to the context of NLP model testing, you should hence communicate the definitions (for the validity or reliability subcomponent) that you work of.}, on which ones are (most) important to test, and even \shortcite<see e.g.,>{borsboom_concept_2004,newton_standards_2013} on whether the concept, as it is currently used, is useful at all. You might thus encounter several different subcomponents of construct validity -- or conceptualizations of validity outside of the construct validity paradigm (e.g., ``criterion validity'') -- that are not mentioned here.
As our goal was to inspire validation research (i.e., research testing whether ``bias measures'' actually measure bias), we chose only the subset of validity conceptualizations that we deemed most conducive towards that goal. 
Some popular subcomponents (like \emph{consequential validity} which concerns the societal impact of widely applying a measurement tool) are not discussed here, as they are unrelated to the question of whether a measurement tool assesses what we want it to. Other subcomponents are related to that question but have substantial overlap with subcomponents we discussed here and hence do not inspire sufficiently different validation efforts to merit extensive discussion.

For example,
concurrent validity and predictive validity \shortcite{cronbach_construct_1955} have conceptual overlap with convergent and divergent validity: Concurrent and predictive validity concern convergent and divergent evidence, but -- instead of emphasizing the nature of measures' relationships (i.e., whether there is convergence or divergence) -- emphasize the timing of measurements: If the comparison measurement is obtained simultaneously with the test we seek to validate, we assess \emph{concurrent validity}; if the measurement occurs after the test we seek to validate (e.g., a kid's math aptitude score is positively correlated with later job performance, but not with later beauty), we assess \emph{predictive validity}. As concurrent and predictive validity make similar prescriptions for validation efforts as do convergent and divergent validity (i.e., ``expect strong positive correlations for convergent relationships and the absence of such correlations for divergent relationships'') and as we consider the ``convergence vs divergence'' distinction more theoretically insightful for NLP bias measures than the ``measured simultaneously vs measured apart'' distinction, we only discussed convergent and divergent validity, here.

\section{From Theory to Practice: Designing Good Bias Measures}
\label{sec:generalizing-bias}

How do we put the lesson from psychometrics into practice when designing bias measures? 
In the following, we present questions and considerations informed by psychometrics, as they apply to three different phases of the bias measure development cycle:
(i) \textit{the preparatory phase} before designing the measure, 
(ii) \textit{the development phase}, where the reliability and construct validity is evaluated, and 
(iii) \textit{the post-development phase} in which results and limitations are communicated.
Our list of questions is not intended to be exhaustive (nor do all have to be answered necessarily); it just provides some examples for the types of issues researchers should consider when developing such a measure, and should help with making some of their assumptions explicit. In that, it should be seen as complementing other guidelines from the literature \shortcite<e.g.,>{blodgett_language_2020,blodgett_stereotyping_2021,dev_measures_2022,talat_you_2022}. 

When considering these questions, keep in mind two things: Firstly, not all of these questions will readily apply to every bias measurement application. Secondly, it is fine to provide answers with low confidence or conviction. More transparency (also if it comes in the form of ``The choice felt unimportant to me, so I picked the easier option'') is always welcome.

\subsection*{(i) Preparation Phase: Understanding the Task and Sociotechnical Context}
The preparation phase occurs before the creation of a bias measure. It involves understanding the desired goals, the task at hand %
and formulating as well as considering the consequences of %
one's assumptions. %

\hfill \break
\textit{Goal Formulation}

\begin{itemize}
     \item What are the relevant forms of bias to measure? Which downstream harms do we want our measure to be predictive of (\S \ref{subsec:why_measure_bias})?
    \item For which (downstream) tasks do we develop this bias measure (\S \ref{subsec:why_measure_bias})? 
    \item %
    What would an ideal bias score be, according to the measure? %
    What does it mean for a bias score to be `low'; and who makes that judgement?
\end{itemize}

\hfill \break
\textit{Preparing Bias Measure Development}
\begin{itemize}
    \item To what linguistic and cultural context(s) do we wish to apply the bias measure?
    \item Does a reliable (\S \ref{sec:assess_reliability}) and valid (\S \ref{sec:assess_validity}) bias measure for this context and task already exist? If so: What does our bias measure add? If not: To what extent can we build on existing measures designed for other bias types or contexts?
    \item What kinds of NLP models %
    will our bias measure be applied to (e.g., autoregressive or masked language models)? What constraints does this imply for our measure (see e.g., Table \ref{tab:example-bias-measures}, WinoBias)? %

\end{itemize}

\hfill \break
\textit{Preparing Validation Efforts}
\begin{itemize}
    \item How many computational resources do we have access to, for our validation efforts (e.g., is it viable to test the seed-based test-retest reliability (\S \ref{subsec:rel_test_retest}) of our measure)?
    \item What are our assumptions about the bias construct (\S \ref{subsec:construct_vs_operationalization})? What (potential) subcomponents of model bias are relevant for this bias measure (\S \ref{subsec:content_validity})? Which constructs do we assume to be related and unrelated to our construct of interest? These can later be used to assess convergent (\S\ref{subsec:convergent-validity}) and divergent validity (\S\ref{subsec:divergent_validity}).
    \item If our measure depends on downstream task performance (e.g., WinoBias, see Table~ \ref{tab:example-bias-measures}): How do we expect the model's bias to influence its behavior on the downstream task (\S \ref{subsec:divergent_validity})? %
    \item What assumptions do we make during the ``translation step'' (\S \ref{subsec:translation_step}) of psychometric concepts to the NLP context? What are the theoretical consequences of making these assumptions? What definitions of reliability (\S \ref{sec:assess_reliability}) and validity (\S \ref{sec:assess_validity}) do we work with?
    \item How can our theoretical assumptions and decision making processes be documented, for later transparency?
\end{itemize}

\subsection*{(ii) Development Phase: Assessing the Reliability and Construct Validity}
Once a first draft of a bias measure has been designed, iterative improvements of the measure can be informed by evaluations of its reliability (\S \ref{sec:assess_reliability}) and validity (\S \ref{sec:assess_validity}). %
Even if considerations about its reliability and validity play no role in the development of a measure%
, at least the reliability and validity of the final bias measure should be evaluated. %

\hfill \break
\textit{Reliability Assessments}
\begin{itemize}
    \item How do we source or generate candidate items for our bias measure? Do alternative formulations result in a similar bias score (\S \ref{subsec:rel_parallel_form})?
    \item Do human annotators judge candidate items for our bias measure? If so, what is the inter-rater reliability of their ratings (\S \ref{subsec:rel_inter_rater})? How robust is our bias score to incorrect annotations?

    \item Are scores on individual items aggregated to produce a total score that our bias measure assigns to an embedding/language model? If so, are model's responses to an individual item of the measure consistent with that overall score (\S \ref{subsec:rel_internal_consistency})? How robust is the measure to removing items of low consistency with the total score? %
    
    \item Is it relevant and feasible to retrain a model to assess the bias measure's \emph{seed-based test-retest reliability} (\S \ref{subsec:rel_test_retest})? Can we assess the bias scores of a model repeatedly, during training? If so, are bias scores largely consistent across proximal training steps?

\end{itemize}

\hfill \break
\textit{Validity Assessments}
\begin{itemize}

    \item Is it feasible to train models which differ in the degree of bias they possess? Does the bias score reflect this as one would expect? %

    \item Does our measure correlate strongly with the (previously identified) important downstream harm(s) (\S\ref{subsec:convergent-validity})? Can we obtain behavioral experimental or survey data of stakeholders, for assessing downstream harms? Are there changes we can make to the measure (e.g., delete test items) to increase these correlations with important downstream harms?\footnote{Make sure to use techniques like cross-validation to ensure that you are not \emph{overfitting} (i.e., optimizing your measure for this particular set of models in a way that does not generalize to other models).} 
    
    \item Do scores of our measure correlate strongly with scores from other measurement tools that are supposed to measure the same construct as ours (\S\ref{subsec:convergent-validity})? %
    
    \item Are there relevant measures for testing the \emph{divergent validity} (\S\ref{subsec:divergent_validity}), that is: measures of constructs that could be confounded with --- but theoretically should not relate to --- our construct? 
    \item Are there ways of estimating the influence of method effects (\S\ref{subsec:divergent_validity}) on our observed correlations (e.g., to judge whether a small positive correlation between measures of uncorrelated constructs implies poor divergent validity or is to be expected, due to method effects)?

    \item Could our measure accommodate subcategorizations of bias (\S \ref{subsec:content_validity})? Does our measure assess all subcomponents of bias that we previously identified as relevant? Did we avoid including construct-irrelevant content in our measure? %

\end{itemize}

\hfill \break
\textit{Practical Considerations}
\begin{itemize}
    \item Given the types of validity and reliability assessments that would theoretically be relevant to our measure, which ones can we implement in practice (due to e.g., computational resources, access to training data)?
    
    \item Could evidence that we deem practically unobtainable be easier to obtain, in the near future? Would it be obtainable with more resources? It is good practice to communicate answers to these questions, during the post-development phase.

    \item How can we facilitate future (re-)evaluations of our measure (e.g., when a new type of relevant downstream harm or other data for establishing convergent validity becomes available)? Do we provide sufficient access (e.g., to training data) and is our record keeping sufficiently precise to enable people outside our research group to perform these (re-)evaluations?
\end{itemize}

\subsection*{(iii) Post-Development Phase: Communicating the Results and Limitations}

In our discussion of the previous phases, we discussed several pieces of information that are important to communicate during the post-developmental phase (e.g., in the Preparation phase: our assumptions about the construct and about the ``translation step''; in the Development phase: how practical considerations influenced our validation efforts, which subcomponents of bias are less well addressed, etc.). Additionally, it is important to be transparent about the following:

\begin{itemize}
    \item For what contexts does the validation assessment hold, and when do we need to perform a new reliability and validity assessment? In other words: Which interpretation of the bias scores was validated, and what should the bias measure not be used for?

    \item Did we reach acceptable levels of reliability and validity? What limitations of the bias measure must be communicated to stakeholders (e.g., which downstream harms are not well-predicted from this measure)? How do these limitations affect the decisions that can be made, based on the measure, about the tested models? %

\end{itemize}

\section{Related Work}
\label{sec:related-works}
We are not the first to discuss validity and reliability concerns of existing bias measures. %
In their survey of bias research in NLP, \shortciteauthor{blodgett_language_2020} \citeyear{blodgett_language_2020} concluded that what researchers meant with \emph{bias} was often poorly defined and inconsistent with the pronounced research goals of the field.
The authors argued for more transparency and proposed that researchers explicitly ground bias measures in the downstream harms of NLP systems (as we also discuss in Section \ref{subsec:why_measure_bias} and \ref{subsec:convergent-validity}). %
Another survey by \shortciteauthor{blodgett_stereotyping_2021} \citeyear{blodgett_stereotyping_2021} --- but of measurement tools based on constrastive sets such as CrowS-Pairs \shortcite{nangia_crows-pairs_2020} and StereoSet \shortcite{nadeem_stereoset_2021} --- categorized an extensive set of examples of bad operationalizations, which threaten the construct validity of these bias measurement benchmarks.
\shortciteauthor{antoniak2021bad} \citeyear{antoniak2021bad} provide (based on experiments and a survey of the literature) a list of factors leading to unreliable results for wordlist-based bias measures.
\shortciteauthor{dev_measures_2022} \citeyear{dev_measures_2022} propose a comprehensive set of questions for improving the documentation of bias measures, including questions concerning the validity.
Similarly to us in advocating for more transparency about one's theoretical assumptions, \shortciteauthor{goldfarb2023prompt} \citeyear{goldfarb2023prompt} surveyed papers of prompt-based bias measures and assessed the extent to which assumptions about the construct were stated and construct and operationalization were consistent with each other.

Few works have, like us, proposed comprehensive frameworks for assessing the construct validity and/or reliability of bias measures. %
However, %
some noteworthy works apply many of the same concepts when evaluating bias measures.
For instance, there are good examples of the types of reliability evaluations that we advocate for in Section \ref{sec:assess_reliability}, with extensive evaluations of the reliability of various static word embedding bias measures \shortcite<see e.g.,>{antoniak2021bad,du_assessing_2021,zhang_robustness_2020}.
More recently, \shortciteauthor{bommasani2022trustworthy}~\citeyear{bommasani2022trustworthy} have applied the framework proposed by \shortciteauthor{jacobs_measurement_2021} \citeyear{jacobs_measurement_2021} for evaluating both the construct validity and reliability of several word embeddings bias measures.

Other noteworthy examples of attempts at ``translating'' psychometric concepts to the NLP context are the works by \shortciteauthor{abbasi2021}~\citeyear{abbasi2021} and \shortciteauthor{fang_evaluating_2022}~\citeyear{fang_evaluating_2022}.
They provide a comprehensive discussion of how to operationalize the constructs of interest and strategies for validating these measures.
However, these two works focus on validating word embedding models for measuring constructs in human-written texts rather than on validating measures to assess a model's internal bias. %

In the related field of algorithmic fairness, other works have emphasized the importance of making a distinction between constructs and their operationalizations \shortcite<e.g.,>{friedler_impossibility_2021,jacobs_measurement_2021}.
Perhaps closest to our work, is the one of \shortciteauthor{jacobs_measurement_2021} \citeyear{jacobs_measurement_2021}, who, similar to our paper, introduce key concepts from psychometrics, including a discussion of types of reliability and construct validity that could be relevant for computational scientists.
However, their focus on measuring fairness in algorithmic decision-making differs from our focus on measuring model bias. As a result, we discuss different methodologies, open questions, and arrive at other recommendations.

To sum up, there is presently a lot of activity on the general topic of the paper, and during the time we have been working on the manuscript many publications came out that addressed very similar issues and worked towards comparable goals. What distinguishes our work from these related efforts is (i) the generality with which the application of the validity and reliability to NLP bias measures is discussed, and (ii) the extent to which background from psychometrics is supplied.

\section{Conclusions}
\label{sec:Discussion}

Bias in NLP is an complex phenomenon, due to its sociotechnical and context-sensitive nature \shortcite{blodgett_language_2020,talat_you_2022}. As a result, researchers face many challenges in the development of measurement and mitigation tools.
In this paper, we addressed the question of how we can test the quality of bias measures, despite these complexities.
In our view, part of the answer is to make use of vocabulary and methodology from psychometrics. Psychological measurements share some of the same challenges as NLP bias (e.g., unobservability of the construct, disagreements between researchers about what ought to be measured). Consequently, their ways of addressing these challenges (e.g., frameworks for assessing reliability and validity) might prove valuable to NLP, as well.

Besides the direct benefits this knowledge transfer will have for the quality of bias measures in NLP, we see also another advantage of building on psychometrics. Its vocabulary will aid NLP researchers to be more transparent and explicit about their conceptualisations of bias and the assumptions they make with their bias measures. This will improve the communication between researchers by helping to contextualize findings (e.g., as pertaining to a particular operationalization versus a particular construct) and by specifying possible points of theoretical convergence and divergence. Note that the benefits of having this vocabulary apply, regardless of whether `gender bias' or any other human-defined type of bias should really be considered one unified thing, or the aggregate of many distinct phenomena. In fact, the distinction between constructs and measures, and between validity and reliability, will also be crucial in any debates about the appropriate level of granularity.

Ultimately, we hope that this better communication and transparency will lead to faster progress in the development of bias measurement tools for embedding and language models. %
Of course, the use of a psychometric lens has its limits. Not all methodologies and insights from the field (readily) apply to the NLP setting. For instance, methodologies designed for human test-takers may be unsuitable for assessing language models (e.g., because we need too many ``test-takers''), or the analogy between a model and a person might break down (e.g., a language model is not subject to time in the same way as people are).
So, adopting a psychometric framework will not solve all issues; there will likely be a need for developing tools specifically for the NLP context.

There is another sense in which the psychometric approach we advocated here is limited.
Designing good measurement tools requires a thorough understanding of the sociocultural context in which the tool is applied. This is particularly pressing in case we measure a complex phenomenon like bias, with all its cultural and sociological connotations. 
To reach this understanding there is a great need for involving other experts (e.g., social scientists, psychologists, philosophers, and linguists) and stakeholders (e.g., designers, owners, and users of these NLP systems, and those potentially harmed by its implementation) in the measurement tool design process \shortcite<see also>[i.a.]{bender_dangers_2021,blodgett_language_2020,dev_measures_2022,kiritchenko_confronting_2021,talat_you_2022}. %

To highlight just one dimension of context dependence, bias measures are bound to the particular language they have been developed for. The fact that a measure is valid or reliable in one linguistic context does not warrant that it transfers well to a different language. Indeed, the bias evaluation of NLP technologies in the multilingual and multicultural setting is especially prone to validity issues \shortcite{blodgett_stereotyping_2021,malik_socially_2022,talat_you_2022}. 
Moreover, bias mitigation efforts do not necessarily transfer between languages even within the same multilingual model \shortcite{gonen_analyzing_2022}. %
These issues are particularly problematic, considering that most research on bias in NLP is focused on one type of bias in one language: gender bias for the English language \shortcite{field_survey_2021,talat_you_2022}. Much more effort needs to be invested in developing proper bias measures for other languages and cultural contexts, keeping in mind that in these other contexts bias might manifest in very different ways than in English \shortcite{ciora2021turkishtranslationgenderbias,jiao_gender_2021}. %

Also in the context of multidisciplinary collaborations and involvements of stakeholders, transparency is key. To give an example, as a society we need to make (normative) choices about where the responsibility of an NLP practitioner ends and where other experts or stakeholders should be involved. A first step towards identifying questions that stakeholders should weigh in on could be to identify disagreements that currently exist within the field of bias measurement --- especially those that do not have empirical answers. Such disagreements can, however, only be unearthed if researchers are explicit about the assumptions they make. 
We hope that our discussion of different types of assumptions (e.g., about construct or operationalization) %
will help NLP researchers refine and communicate their individual understandings of bias --- mitigating the current conceptual confusion \cite{blodgett_language_2020,dev_measures_2022}.

Like \shortciteauthor{jacobs_measurement_2021} \citeyear{jacobs_measurement_2021}, we are only an early effort towards applying measurement theory and psychometric concepts to AI.  As such, we do not want to imply that our perspectives on the topic are definite or gospel. Instead, we hope that we further opened the door towards applying psychometric concepts to AI and invite theoretical discussions of their merits (or conversely, inapplicability) to NLP bias research.

\newpage
\acks{
The two first authors, Oskar van der Wal and Dominik Bachmann, contributed equally to the paper. We thank our four anonymous reviewers for their thoughtful and elaborate feedback. %

This publication is part of the project ``The biased reality of online media - Using stereotypes to make media manipulation visible'' (with project number 406.DI.19.059) of the research programme Open Competition Digitalisation-SSH, which is financed by the Dutch Research Council (NWO).}

\vskip 0.2in
\DeclareRobustCommand{\VAN}[3]{#3}
\bibliography{zotero_references,references}

\begin{thebibliography}{}

\bibitem[\protect\BCAY{Abbasi, Dobolyi, Lalor, Netemeyer, Smith,\ \BBA\
  Yang}{Abbasi et~al.}{2021}]{abbasi2021}
Abbasi, A., Dobolyi, D., Lalor, J.~P., Netemeyer, R.~G., Smith, K., \BBA\ Yang,
  Y. \BBOP2021\BBCP.
\newblock \BBOQ Constructing a {Psychometric} {Testbed} for {Fair} {Natural}
  {Language} {Processing}\BBCQ\
\newblock In {\Bem Proceedings of the 2021 {Conference} on {Empirical}
  {Methods} in {Natural} {Language} {Processing}}, \BPGS\ 3748--3758, Online
  and Punta Cana, Dominican Republic. Association for Computational
  Linguistics.

\bibitem[\protect\BCAY{Aky{\"u}rek, Paik, Kocyigit, Akbiyik, Runyun,\ \BBA\
  Wijaya}{Aky{\"u}rek et~al.}{2022}]{akyurek2022measuring}
Aky{\"u}rek, A.~F., Paik, S., Kocyigit, M., Akbiyik, S., Runyun, S.~L., \BBA\
  Wijaya, D. \BBOP2022\BBCP.
\newblock \BBOQ On measuring social biases in prompt-based multi-task
  learning\BBCQ\
\newblock In {\Bem Findings of the Association for Computational Linguistics:
  NAACL 2022}, \BPGS\ 551--564.

\bibitem[\protect\BCAY{Akyürek, Kocyigit, Paik,\ \BBA\ Wijaya}{Akyürek
  et~al.}{2022}]{akyurek_challenges_2022}
Akyürek, A.~F., Kocyigit, M.~Y., Paik, S., \BBA\ Wijaya, D.~T. \BBOP2022\BBCP.
\newblock \BBOQ Challenges in {Measuring} {Bias} via {Open}-{Ended} {Language}
  {Generation}\BBCQ\
\newblock In {\Bem Proceedings of the 4th {Workshop} on {Gender} {Bias} in
  {Natural} {Language} {Processing} ({GeBNLP})}, \BPGS\ 76--76, Seattle,
  Washington. Association for Computational Linguistics.

\bibitem[\protect\BCAY{Amidei, Piwek,\ \BBA\ Willis}{Amidei
  et~al.}{2020}]{amidei_identifying_2020}
Amidei, J., Piwek, P., \BBA\ Willis, A. \BBOP2020\BBCP.
\newblock \BBOQ Identifying {Annotator} {Bias}: {A} new {IRT}-based method for
  bias identification\BBCQ\
\newblock In {\Bem Proceedings of the 28th {International} {Conference} on
  {Computational} {Linguistics}}, \BPGS\ 4787--4797, Barcelona, Spain (Online).
  International Committee on Computational Linguistics.

\bibitem[\protect\BCAY{Antoniak\ \BBA\ Mimno}{Antoniak\ \BBA\
  Mimno}{2021}]{antoniak2021bad}
Antoniak, M.\BBACOMMA\  \BBA\ Mimno, D. \BBOP2021\BBCP.
\newblock \BBOQ Bad seeds: Evaluating lexical methods for bias
  measurement\BBCQ\
\newblock In {\Bem Proceedings of the 59th Annual Meeting of the Association
  for Computational Linguistics and the 11th International Joint Conference on
  Natural Language Processing (Volume 1: Long Papers)}, \BPGS\ 1889--1904.

\bibitem[\protect\BCAY{Aribandi, Tay,\ \BBA\ Metzler}{Aribandi
  et~al.}{2021}]{aribandi_how_2021}
Aribandi, V., Tay, Y., \BBA\ Metzler, D. \BBOP2021\BBCP.
\newblock \BBOQ How {Reliable} are {Model} {Diagnostics}?\BBCQ\
\newblock In {\Bem Findings of the {Association} for {Computational}
  {Linguistics}: {ACL}-{IJCNLP} 2021}, \BPGS\ 1778--1785, Online. Association
  for Computational Linguistics.

\bibitem[\protect\BCAY{Artstein\ \BBA\ Poesio}{Artstein\ \BBA\
  Poesio}{2008}]{artstein_inter-coder_2008}
Artstein, R.\BBACOMMA\  \BBA\ Poesio, M. \BBOP2008\BBCP.
\newblock \BBOQ Inter-{Coder} {Agreement} for {Computational}
  {Linguistics}\BBCQ\
\newblock {\Bem Computational Linguistics}, {\Bem 34\/}(4), 555--596.

\bibitem[\protect\BCAY{Bach, Sanh, Yong, Webson, Raffel, Nayak, Sharma, Kim,
  Bari, F{\'e}vry, et~al.}{Bach et~al.}{2022}]{bach2022promptsource}
Bach, S., Sanh, V., Yong, Z.~X., Webson, A., Raffel, C., Nayak, N.~V., Sharma,
  A., Kim, T., Bari, M.~S., F{\'e}vry, T., et~al. \BBOP2022\BBCP.
\newblock \BBOQ Promptsource: An integrated development environment and
  repository for natural language prompts\BBCQ\
\newblock In {\Bem Proceedings of the 60th Annual Meeting of the Association
  for Computational Linguistics: System Demonstrations}, \BPGS\ 93--104.

\bibitem[\protect\BCAY{Balayn\ \BBA\ G{\"u}rses}{Balayn\ \BBA\
  G{\"u}rses}{2021}]{balayn2021beyond}
Balayn, A.\BBACOMMA\  \BBA\ G{\"u}rses, S. \BBOP2021\BBCP.
\newblock \BBOQ Beyond debiasing: Regulating ai and its inequalities\BBCQ\
\newblock {\Bem EDRi Report}.

\bibitem[\protect\BCAY{Barocas, Crawford, Shapiro,\ \BBA\ Wallach}{Barocas
  et~al.}{2017}]{barocas2017problem}
Barocas, S., Crawford, K., Shapiro, A., \BBA\ Wallach, H. \BBOP2017\BBCP.
\newblock \BBOQ The problem with bias: Allocative versus representational harms
  in machine learning\BBCQ\
\newblock In {\Bem 9th Annual conference of the special interest group for
  computing, information and society}.

\bibitem[\protect\BCAY{Bender, Gebru, McMillan-Major,\ \BBA\ Shmitchell}{Bender
  et~al.}{2021}]{bender_dangers_2021}
Bender, E.~M., Gebru, T., McMillan-Major, A., \BBA\ Shmitchell, S.
  \BBOP2021\BBCP.
\newblock \BBOQ On the {Dangers} of {Stochastic} {Parrots}: {Can} {Language}
  {Models} {Be} {Too} {Big}? 🦜\BBCQ\
\newblock In {\Bem Proceedings of the 2021 {ACM} {Conference} on {Fairness},
  {Accountability}, and {Transparency}}, {FAccT} '21, \BPGS\ 610--623, New
  York, NY, USA. Association for Computing Machinery.
\newblock tex.ids= bender2021DangersStochasticParrotsa.

\bibitem[\protect\BCAY{Bhowmick, Basu,\ \BBA\ Mitra}{Bhowmick
  et~al.}{2008}]{bhowmick_agreement_2008}
Bhowmick, P.~K., Basu, A., \BBA\ Mitra, P. \BBOP2008\BBCP.
\newblock \BBOQ An {Agreement} {Measure} for {Determining} {Inter}-{Annotator}
  {Reliability} of {Human} {Judgements} on {Affective} {Text}\BBCQ\
\newblock In {\Bem Coling 2008: {Proceedings} of the workshop on {Human}
  {Judgements} in {Computational} {Linguistics}}, \BPGS\ 58--65, Manchester,
  UK. Coling 2008 Organizing Committee.

\bibitem[\protect\BCAY{Biderman, Schoelkopf, Anthony, Bradley, O'Brien,
  Hallahan, Khan, Purohit, Prashanth, Raff, Skowron, Sutawika,\ \BBA\ van~der
  Wal}{Biderman et~al.}{2023}]{biderman2023pythia}
Biderman, S., Schoelkopf, H., Anthony, Q.~G., Bradley, H., O'Brien, K.,
  Hallahan, E., Khan, M.~A., Purohit, S., Prashanth, U.~S., Raff, E., Skowron,
  A., Sutawika, L., \BBA\ van~der Wal, O. \BBOP2023\BBCP.
\newblock \BBOQ Pythia: {A} suite for analyzing large language models across
  training and scaling\BBCQ\
\newblock In {\Bem International Conference on Machine Learning, {ICML} 2023,
  23-29 July 2023, Honolulu, Hawaii, {USA}}, \lowercase{\BVOL}\ 202 of {\Bem
  Proceedings of Machine Learning Research}, \BPGS\ 2397--2430. {PMLR}.

\bibitem[\protect\BCAY{Blodgett, Barocas, Daumé~III,\ \BBA\ Wallach}{Blodgett
  et~al.}{2020}]{blodgett_language_2020}
Blodgett, S.~L., Barocas, S., Daumé~III, H., \BBA\ Wallach, H. \BBOP2020\BBCP.
\newblock \BBOQ Language ({Technology}) is {Power}: {A} {Critical} {Survey} of
  “{Bias}” in {NLP}\BBCQ\
\newblock In {\Bem Proceedings of the 58th {Annual} {Meeting} of the
  {Association} for {Computational} {Linguistics}}, \BPGS\ 5454--5476, Online.
  Association for Computational Linguistics.

\bibitem[\protect\BCAY{Blodgett, Lopez, Olteanu, Sim,\ \BBA\ Wallach}{Blodgett
  et~al.}{2021}]{blodgett_stereotyping_2021}
Blodgett, S.~L., Lopez, G., Olteanu, A., Sim, R., \BBA\ Wallach, H.
  \BBOP2021\BBCP.
\newblock \BBOQ Stereotyping {Norwegian} {Salmon}: {An} {Inventory} of
  {Pitfalls} in {Fairness} {Benchmark} {Datasets}\BBCQ\
\newblock In {\Bem Proceedings of the 59th {Annual} {Meeting} of the
  {Association} for {Computational} {Linguistics} and the 11th {International}
  {Joint} {Conference} on {Natural} {Language} {Processing} ({Volume} 1: {Long}
  {Papers})}, \BPGS\ 1004--1015, Online. Association for Computational
  Linguistics.

\bibitem[\protect\BCAY{Bolukbasi, Chang, Zou, Saligrama,\ \BBA\
  Kalai}{Bolukbasi et~al.}{2016}]{bolukbasi_man_2016}
Bolukbasi, T., Chang, K.-W., Zou, J., Saligrama, V., \BBA\ Kalai, A.
  \BBOP2016\BBCP.
\newblock \BBOQ Man is to computer programmer as woman is to homemaker?
  debiasing word embeddings\BBCQ\
\newblock In {\Bem Proceedings of the 30th {International} {Conference} on
  {Neural} {Information} {Processing} {Systems}}, {NIPS}'16, \BPGS\ 4356--4364,
  Red Hook, NY, USA. Curran Associates Inc.

\bibitem[\protect\BCAY{Bommasani, Hudson, Adeli, Altman, Arora, von Arx,
  Bernstein, Bohg, Bosselut, Brunskill, et~al.}{Bommasani
  et~al.}{2021}]{bommasani_opportunities_2022b}
Bommasani, R., Hudson, D.~A., Adeli, E., Altman, R., Arora, S., von Arx, S.,
  Bernstein, M.~S., Bohg, J., Bosselut, A., Brunskill, E., et~al.
  \BBOP2021\BBCP.
\newblock \BBOQ On the opportunities and risks of foundation models\BBCQ\
\newblock {\Bem arXiv preprint arXiv:2108.07258}.

\bibitem[\protect\BCAY{Bommasani\ \BBA\ Liang}{Bommasani\ \BBA\
  Liang}{2022}]{bommasani2022trustworthy}
Bommasani, R.\BBACOMMA\  \BBA\ Liang, P. \BBOP2022\BBCP.
\newblock \BBOQ Trustworthy social bias measurement\BBCQ\
\newblock {\Bem arXiv preprint arXiv:2212.11672}.

\bibitem[\protect\BCAY{Bordia\ \BBA\ Bowman}{Bordia\ \BBA\
  Bowman}{2019}]{bordia_identifying_2019}
Bordia, S.\BBACOMMA\  \BBA\ Bowman, S.~R. \BBOP2019\BBCP.
\newblock \BBOQ Identifying and {Reducing} {Gender} {Bias} in {Word}-{Level}
  {Language} {Models}\BBCQ\
\newblock In {\Bem Proceedings of the 2019 {Conference} of the {North}
  {American} {Chapter} of the {Association} for {Computational} {Linguistics}:
  {Student} {Research} {Workshop}}, \BPGS\ 7--15, Minneapolis, Minnesota.
  Association for Computational Linguistics.

\bibitem[\protect\BCAY{Borsboom, Mellenbergh,\ \BBA\ van Heerden}{Borsboom
  et~al.}{2004}]{borsboom_concept_2004}
Borsboom, D., Mellenbergh, G.~J., \BBA\ van Heerden, J. \BBOP2004\BBCP.
\newblock \BBOQ The {Concept} of {Validity}\BBCQ\
\newblock {\Bem Psychological Review}, {\Bem 111\/}(4), 1061--1071.
\newblock Place: US Publisher: American Psychological Association.

\bibitem[\protect\BCAY{Brown, Mann, Ryder, Subbiah, Kaplan, Dhariwal,
  Neelakantan, Shyam, Sastry, Askell, Agarwal, Herbert-Voss, Krueger, Henighan,
  Child, Ramesh, Ziegler, Wu, Winter, Hesse, Chen, Sigler, Litwin, Gray, Chess,
  Clark, Berner, McCandlish, Radford, Sutskever,\ \BBA\ Amodei}{Brown
  et~al.}{2020}]{brown_language_2020}
Brown, T., Mann, B., Ryder, N., Subbiah, M., Kaplan, J.~D., Dhariwal, P.,
  Neelakantan, A., Shyam, P., Sastry, G., Askell, A., Agarwal, S.,
  Herbert-Voss, A., Krueger, G., Henighan, T., Child, R., Ramesh, A., Ziegler,
  D., Wu, J., Winter, C., Hesse, C., Chen, M., Sigler, E., Litwin, M., Gray,
  S., Chess, B., Clark, J., Berner, C., McCandlish, S., Radford, A., Sutskever,
  I., \BBA\ Amodei, D. \BBOP2020\BBCP.
\newblock \BBOQ Language {Models} are {Few}-{Shot} {Learners}\BBCQ\
\newblock In {\Bem Advances in {Neural} {Information} {Processing} {Systems}},
  \lowercase{\BVOL}~33, \BPGS\ 1877--1901. Curran Associates, Inc.

\bibitem[\protect\BCAY{Caliskan, Bryson,\ \BBA\ Narayanan}{Caliskan
  et~al.}{2017}]{caliskanSemanticsDerivedAutomatically2017}
Caliskan, A., Bryson, J.~J., \BBA\ Narayanan, A. \BBOP2017\BBCP.
\newblock \BBOQ Semantics derived automatically from language corpora contain
  human-like biases\BBCQ\
\newblock {\Bem Science}, {\Bem 356\/}(6334), 183--186.

\bibitem[\protect\BCAY{Campbell\ \BBA\ Fiske}{Campbell\ \BBA\
  Fiske}{1959}]{campbell1959convergent}
Campbell, D.~T.\BBACOMMA\  \BBA\ Fiske, D.~W. \BBOP1959\BBCP.
\newblock \BBOQ Convergent and discriminant validation by the
  multitrait-multimethod matrix\BBCQ\
\newblock {\Bem Psychological bulletin}, {\Bem 56\/}(2), 81--105.

\bibitem[\protect\BCAY{Cao, Pruksachatkun, Chang, Gupta, Kumar, Dhamala,\ \BBA\
  Galstyan}{Cao et~al.}{2022}]{cao_intrinsic_2022}
Cao, Y., Pruksachatkun, Y., Chang, K.-W., Gupta, R., Kumar, V., Dhamala, J.,
  \BBA\ Galstyan, A. \BBOP2022\BBCP.
\newblock \BBOQ On the {Intrinsic} and {Extrinsic} {Fairness} {Evaluation}
  {Metrics} for {Contextualized} {Language} {Representations}\BBCQ\
\newblock In {\Bem Proceedings of the 60th {Annual} {Meeting} of the
  {Association} for {Computational} {Linguistics} ({Volume} 2: {Short}
  {Papers})}, \BPGS\ 561--570, Dublin, Ireland. Association for Computational
  Linguistics.

\bibitem[\protect\BCAY{Cheng, Varshney,\ \BBA\ Liu}{Cheng
  et~al.}{2021}]{cheng_socially_2021}
Cheng, L., Varshney, K.~R., \BBA\ Liu, H. \BBOP2021\BBCP.
\newblock \BBOQ Socially {Responsible} {AI} {Algorithms}: {Issues}, {Purposes},
  and {Challenges}\BBCQ\
\newblock {\Bem Journal of Artificial Intelligence Research}, {\Bem 71},
  1137--1181.

\bibitem[\protect\BCAY{Ciora, Iren,\ \BBA\ Alikhani}{Ciora
  et~al.}{2021}]{ciora2021turkishtranslationgenderbias}
Ciora, C., Iren, N., \BBA\ Alikhani, M. \BBOP2021\BBCP.
\newblock \BBOQ Examining covert gender bias: A case study in turkish and
  english machine translation models\BBCQ\
\newblock In {\Bem Proceedings of the 14th International Conference on Natural
  Language Generation}, \BPGS\ 55--63.

\bibitem[\protect\BCAY{Cronbach\ \BBA\ Meehl}{Cronbach\ \BBA\
  Meehl}{1955}]{cronbach_construct_1955}
Cronbach, L.~J.\BBACOMMA\  \BBA\ Meehl, P.~E. \BBOP1955\BBCP.
\newblock \BBOQ Construct validity in psychological tests\BBCQ\
\newblock {\Bem Psychological Bulletin}, {\Bem 52\/}(4), 281--302.
\newblock Place: US Publisher: American Psychological Association.

\bibitem[\protect\BCAY{D'Amour, Heller, Moldovan, Adlam, Alipanahi, Beutel,
  Chen, Deaton, Eisenstein, Hoffman, Hormozdiari, Houlsby, Hou, Jerfel,
  Karthikesalingam, Lucic, Ma, McLean, Mincu, Mitani, Montanari, Nado,
  Natarajan, Nielson, Osborne, Raman, Ramasamy, Sayres, Schrouff, Seneviratne,
  Sequeira, Suresh, Veitch, Vladymyrov, Wang, Webster, Yadlowsky, Yun, Zhai,\
  \BBA\ Sculley}{D'Amour et~al.}{2022}]{damour_underspecification_2022}
D'Amour, A., Heller, K., Moldovan, D., Adlam, B., Alipanahi, B., Beutel, A.,
  Chen, C., Deaton, J., Eisenstein, J., Hoffman, M.~D., Hormozdiari, F.,
  Houlsby, N., Hou, S., Jerfel, G., Karthikesalingam, A., Lucic, M., Ma, Y.,
  McLean, C., Mincu, D., Mitani, A., Montanari, A., Nado, Z., Natarajan, V.,
  Nielson, C., Osborne, T.~F., Raman, R., Ramasamy, K., Sayres, R., Schrouff,
  J., Seneviratne, M., Sequeira, S., Suresh, H., Veitch, V., Vladymyrov, M.,
  Wang, X., Webster, K., Yadlowsky, S., Yun, T., Zhai, X., \BBA\ Sculley, D.
  \BBOP2022\BBCP.
\newblock \BBOQ Underspecification {Presents} {Challenges} for {Credibility} in
  {Modern} {Machine} {Learning}\BBCQ\
\newblock {\Bem Journal of Machine Learning Research}, {\Bem 23\/}(226), 1--61.

\bibitem[\protect\BCAY{Danesi}{Danesi}{2014}]{danesi2014dictionary}
Danesi, M. \BBOP2014\BBCP.
\newblock {\Bem Dictionary of media and communications}.
\newblock Routledge.

\bibitem[\protect\BCAY{De-Arteaga, Romanov, Wallach, Chayes, Borgs,
  Chouldechova, Geyik, Kenthapadi,\ \BBA\ Kalai}{De-Arteaga
  et~al.}{2019}]{de-arteaga_bias_2019}
De-Arteaga, M., Romanov, A., Wallach, H., Chayes, J., Borgs, C., Chouldechova,
  A., Geyik, S., Kenthapadi, K., \BBA\ Kalai, A.~T. \BBOP2019\BBCP.
\newblock \BBOQ Bias in {Bios}: {A} {Case} {Study} of {Semantic}
  {Representation} {Bias} in a {High}-{Stakes} {Setting}\BBCQ\
\newblock In {\Bem Proceedings of the {Conference} on {Fairness},
  {Accountability}, and {Transparency}}, {FAT}* '19, \BPGS\ 120--128, New York,
  NY, USA. Association for Computing Machinery.

\bibitem[\protect\BCAY{De~Cao, Schmid, Hupkes,\ \BBA\ Titov}{De~Cao
  et~al.}{2022}]{de_cao_sparse_2022}
De~Cao, N., Schmid, L., Hupkes, D., \BBA\ Titov, I. \BBOP2022\BBCP.
\newblock \BBOQ Sparse {Interventions} in {Language} {Models} with
  {Differentiable} {Masking}\BBCQ\
\newblock In {\Bem Proceedings of the {Fifth} {BlackboxNLP} {Workshop} on
  {Analyzing} and {Interpreting} {Neural} {Networks} for {NLP}}, \BPGS\ 16--27,
  Abu Dhabi, United Arab Emirates (Hybrid). Association for Computational
  Linguistics.

\bibitem[\protect\BCAY{Dehghani, Tay, Gritsenko, Zhao, Houlsby, Diaz, Metzler,\
  \BBA\ Vinyals}{Dehghani et~al.}{2021}]{dehghani2021benchmark}
Dehghani, M., Tay, Y., Gritsenko, A.~A., Zhao, Z., Houlsby, N., Diaz, F.,
  Metzler, D., \BBA\ Vinyals, O. \BBOP2021\BBCP.
\newblock \BBOQ The benchmark lottery\BBCQ\
\newblock {\Bem arXiv preprint arXiv:2107.07002}.

\bibitem[\protect\BCAY{Delobelle, Tokpo, Calders,\ \BBA\ Berendt}{Delobelle
  et~al.}{2022}]{delobelle_measuring_2022}
Delobelle, P., Tokpo, E., Calders, T., \BBA\ Berendt, B. \BBOP2022\BBCP.
\newblock \BBOQ Measuring {Fairness} with {Biased} {Rulers}: {A} {Comparative}
  {Study} on {Bias} {Metrics} for {Pre}-trained {Language} {Models}\BBCQ\
\newblock In {\Bem Proceedings of the 2022 {Conference} of the {North}
  {American} {Chapter} of the {Association} for {Computational} {Linguistics}:
  {Human} {Language} {Technologies}}, \BPGS\ 1693--1706, Seattle, United
  States. Association for Computational Linguistics.

\bibitem[\protect\BCAY{Dev, Monajatipoor, Ovalle, Subramonian, Phillips,\ \BBA\
  Chang}{Dev et~al.}{2021}]{dev_harms_2021}
Dev, S., Monajatipoor, M., Ovalle, A., Subramonian, A., Phillips, J., \BBA\
  Chang, K.-W. \BBOP2021\BBCP.
\newblock \BBOQ Harms of {Gender} {Exclusivity} and {Challenges} in
  {Non}-{Binary} {Representation} in {Language} {Technologies}\BBCQ\
\newblock In {\Bem Proceedings of the 2021 {Conference} on {Empirical}
  {Methods} in {Natural} {Language} {Processing}}, \BPGS\ 1968--1994, Online
  and Punta Cana, Dominican Republic. Association for Computational
  Linguistics.

\bibitem[\protect\BCAY{Dev, Sheng, Zhao, Amstutz, Sun, Hou, Sanseverino, Kim,
  Nishi, Peng,\ \BBA\ Chang}{Dev et~al.}{2022}]{dev_measures_2022}
Dev, S., Sheng, E., Zhao, J., Amstutz, A., Sun, J., Hou, Y., Sanseverino, M.,
  Kim, J., Nishi, A., Peng, N., \BBA\ Chang, K.-W. \BBOP2022\BBCP.
\newblock \BBOQ On {Measures} of {Biases} and {Harms} in {NLP}\BBCQ\
\newblock In {\Bem Findings of the {Association} for {Computational}
  {Linguistics}: {AACL}-{IJCNLP} 2022}, \BPGS\ 246--267, Online only.
  Association for Computational Linguistics.

\bibitem[\protect\BCAY{Dinan, Fan, Wu, Weston, Kiela,\ \BBA\ Williams}{Dinan
  et~al.}{2020}]{dinan_multi-dimensional_2020}
Dinan, E., Fan, A., Wu, L., Weston, J., Kiela, D., \BBA\ Williams, A.
  \BBOP2020\BBCP.
\newblock \BBOQ Multi-{Dimensional} {Gender} {Bias} {Classification}\BBCQ\
\newblock In {\Bem Proceedings of the 2020 {Conference} on {Empirical}
  {Methods} in {Natural} {Language} {Processing} ({EMNLP})}, \BPGS\ 314--331,
  Online. Association for Computational Linguistics.

\bibitem[\protect\BCAY{Du, Fang,\ \BBA\ Nguyen}{Du
  et~al.}{2021}]{du_assessing_2021}
Du, Y., Fang, Q., \BBA\ Nguyen, D. \BBOP2021\BBCP.
\newblock \BBOQ Assessing the {Reliability} of {Word} {Embedding} {Gender}
  {Bias} {Measures}\BBCQ\
\newblock In {\Bem Proceedings of the 2021 {Conference} on {Empirical}
  {Methods} in {Natural} {Language} {Processing}}, \BPGS\ 10012--10034, Online
  and Punta Cana, Dominican Republic. Association for Computational
  Linguistics.

\bibitem[\protect\BCAY{Eagly, Nater, Miller, Kaufmann,\ \BBA\ Sczesny}{Eagly
  et~al.}{2020}]{eagly_gender_2020}
Eagly, A.~H., Nater, C., Miller, D.~I., Kaufmann, M., \BBA\ Sczesny, S.
  \BBOP2020\BBCP.
\newblock \BBOQ Gender stereotypes have changed: {A} cross-temporal
  meta-analysis of {U}.{S}. public opinion polls from 1946 to 2018\BBCQ\
\newblock {\Bem American Psychologist}, {\Bem 75}, 301--315.
\newblock place: US publisher: American Psychological Association.

\bibitem[\protect\BCAY{Ethayarajh}{Ethayarajh}{2020}]{ethayarajh_is_2020}
Ethayarajh, K. \BBOP2020\BBCP.
\newblock \BBOQ Is {Your} {Classifier} {Actually} {Biased}? {Measuring}
  {Fairness} under {Uncertainty} with {Bernstein} {Bounds}\BBCQ\
\newblock In {\Bem Proceedings of the 58th {Annual} {Meeting} of the
  {Association} for {Computational} {Linguistics}}, \BPGS\ 2914--2919, Online.
  Association for Computational Linguistics.

\bibitem[\protect\BCAY{Ethayarajh, Duvenaud,\ \BBA\ Hirst}{Ethayarajh
  et~al.}{2019}]{ethayarajh_understanding_2019}
Ethayarajh, K., Duvenaud, D., \BBA\ Hirst, G. \BBOP2019\BBCP.
\newblock \BBOQ Understanding {Undesirable} {Word} {Embedding}
  {Associations}\BBCQ\
\newblock In {\Bem Proceedings of the 57th {Annual} {Meeting} of the
  {Association} for {Computational} {Linguistics}}, \BPGS\ 1696--1705,
  Florence, Italy. Association for Computational Linguistics.

\bibitem[\protect\BCAY{Fang, Nguyen,\ \BBA\ Oberski}{Fang
  et~al.}{2022}]{fang_evaluating_2022}
Fang, Q., Nguyen, D., \BBA\ Oberski, D.~L. \BBOP2022\BBCP.
\newblock \BBOQ Evaluating the construct validity of text embeddings with
  application to survey questions\BBCQ\
\newblock {\Bem EPJ Data Science}, {\Bem 11\/}(1), 39.

\bibitem[\protect\BCAY{Field, Blodgett, Waseem,\ \BBA\ Tsvetkov}{Field
  et~al.}{2021}]{field_survey_2021}
Field, A., Blodgett, S.~L., Waseem, Z., \BBA\ Tsvetkov, Y. \BBOP2021\BBCP.
\newblock \BBOQ A {Survey} of {Race}, {Racism}, and {Anti}-{Racism} in
  {NLP}\BBCQ\
\newblock In {\Bem Proceedings of the 59th {Annual} {Meeting} of the
  {Association} for {Computational} {Linguistics} and the 11th {International}
  {Joint} {Conference} on {Natural} {Language} {Processing} ({Volume} 1: {Long}
  {Papers})}, \BPGS\ 1905--1925, Online. Association for Computational
  Linguistics.

\bibitem[\protect\BCAY{Field\ \BBA\ Tsvetkov}{Field\ \BBA\
  Tsvetkov}{2020}]{field_unsupervised_2020}
Field, A.\BBACOMMA\  \BBA\ Tsvetkov, Y. \BBOP2020\BBCP.
\newblock \BBOQ Unsupervised {Discovery} of {Implicit} {Gender} {Bias}\BBCQ\
\newblock In {\Bem Proceedings of the 2020 {Conference} on {Empirical}
  {Methods} in {Natural} {Language} {Processing} ({EMNLP})}, \BPGS\ 596--608,
  Online. Association for Computational Linguistics.

\bibitem[\protect\BCAY{Founta, Djouvas, Chatzakou, Leontiadis, Blackburn,
  Stringhini, Vakali, Sirivianos,\ \BBA\ Kourtellis}{Founta
  et~al.}{2018}]{founta2018large}
Founta, A., Djouvas, C., Chatzakou, D., Leontiadis, I., Blackburn, J.,
  Stringhini, G., Vakali, A., Sirivianos, M., \BBA\ Kourtellis, N.
  \BBOP2018\BBCP.
\newblock \BBOQ Large scale crowdsourcing and characterization of twitter
  abusive behavior\BBCQ\
\newblock In {\Bem Proceedings of the international AAAI conference on web and
  social media}, \lowercase{\BVOL}~12.

\bibitem[\protect\BCAY{Friedler, Scheidegger,\ \BBA\
  Venkatasubramanian}{Friedler et~al.}{2021}]{friedler_impossibility_2021}
Friedler, S.~A., Scheidegger, C., \BBA\ Venkatasubramanian, S. \BBOP2021\BBCP.
\newblock \BBOQ The ({Im})possibility of fairness: different value systems
  require different mechanisms for fair decision making\BBCQ\
\newblock {\Bem Communications of the ACM}, {\Bem 64\/}(4), 136--143.

\bibitem[\protect\BCAY{Garg, Schiebinger, Jurafsky,\ \BBA\ Zou}{Garg
  et~al.}{2018}]{garg2018word}
Garg, N., Schiebinger, L., Jurafsky, D., \BBA\ Zou, J. \BBOP2018\BBCP.
\newblock \BBOQ Word embeddings quantify 100 years of gender and ethnic
  stereotypes\BBCQ\
\newblock {\Bem Proceedings of the National Academy of Sciences}, {\Bem
  115\/}(16), E3635--E3644.

\bibitem[\protect\BCAY{Golchin\ \BBA\ Surdeanu}{Golchin\ \BBA\
  Surdeanu}{2023}]{golchin2023time}
Golchin, S.\BBACOMMA\  \BBA\ Surdeanu, M. \BBOP2023\BBCP.
\newblock \BBOQ Time travel in llms: Tracing data contamination in large
  language models\BBCQ\
\newblock {\Bem arXiv preprint arXiv:2308.08493}.

\bibitem[\protect\BCAY{Goldfarb-Tarrant, Marchant, Muñoz~Sánchez, Pandya,\
  \BBA\ Lopez}{Goldfarb-Tarrant et~al.}{2021}]{goldfarb-tarrant_intrinsic_2021}
Goldfarb-Tarrant, S., Marchant, R., Muñoz~Sánchez, R., Pandya, M., \BBA\
  Lopez, A. \BBOP2021\BBCP.
\newblock \BBOQ Intrinsic {Bias} {Metrics} {Do} {Not} {Correlate} with
  {Application} {Bias}\BBCQ\
\newblock In {\Bem Proceedings of the 59th {Annual} {Meeting} of the
  {Association} for {Computational} {Linguistics} and the 11th {International}
  {Joint} {Conference} on {Natural} {Language} {Processing} ({Volume} 1: {Long}
  {Papers})}, \BPGS\ 1926--1940, Online. Association for Computational
  Linguistics.

\bibitem[\protect\BCAY{Goldfarb-Tarrant, Ungless, Balkir,\ \BBA\
  Blodgett}{Goldfarb-Tarrant et~al.}{2023}]{goldfarb2023prompt}
Goldfarb-Tarrant, S., Ungless, E., Balkir, E., \BBA\ Blodgett, S.~L.
  \BBOP2023\BBCP.
\newblock \BBOQ This prompt is measuring $<${MASK}$>$: Evaluating bias
  evaluation in language models\BBCQ\
\newblock {\Bem arXiv preprint arXiv:2305.12757}.

\bibitem[\protect\BCAY{Gonen\ \BBA\ Goldberg}{Gonen\ \BBA\
  Goldberg}{2019}]{gonen_lipstick_2019}
Gonen, H.\BBACOMMA\  \BBA\ Goldberg, Y. \BBOP2019\BBCP.
\newblock \BBOQ Lipstick on a {Pig}: {Debiasing} {Methods} {Cover} up
  {Systematic} {Gender} {Biases} in {Word} {Embeddings} {But} do not {Remove}
  {Them}\BBCQ\
\newblock In {\Bem Proceedings of the 2019 {Conference} of the {North}
  {American} {Chapter} of the {Association} for {Computational} {Linguistics}:
  {Human} {Language} {Technologies}, {Volume} 1 ({Long} and {Short} {Papers})},
  \BPGS\ 609--614, Minneapolis, Minnesota. Association for Computational
  Linguistics.

\bibitem[\protect\BCAY{Gonen, Ravfogel,\ \BBA\ Goldberg}{Gonen
  et~al.}{2022}]{gonen_analyzing_2022}
Gonen, H., Ravfogel, S., \BBA\ Goldberg, Y. \BBOP2022\BBCP.
\newblock \BBOQ Analyzing {Gender} {Representation} in {Multilingual}
  {Models}\BBCQ\
\newblock In {\Bem Proceedings of the 7th {Workshop} on {Representation}
  {Learning} for {NLP}}, \BPGS\ 67--77, Dublin, Ireland. Association for
  Computational Linguistics.

\bibitem[\protect\BCAY{Greenwald, McGhee,\ \BBA\ Schwartz}{Greenwald
  et~al.}{1998}]{greenwald1998}
Greenwald, A.~G., McGhee, D.~E., \BBA\ Schwartz, J.~L. \BBOP1998\BBCP.
\newblock \BBOQ Measuring individual differences in implicit cognition: the
  implicit association test\BBCQ\
\newblock {\Bem Journal of personality and social psychology}, {\Bem 74\/}(6),
  1464.

\bibitem[\protect\BCAY{Greenwald, Poehlman, Uhlmann,\ \BBA\ Banaji}{Greenwald
  et~al.}{2009}]{greenwald_understanding_2009}
Greenwald, A.~G., Poehlman, T.~A., Uhlmann, E.~L., \BBA\ Banaji, M.~R.
  \BBOP2009\BBCP.
\newblock \BBOQ Understanding and using the {Implicit} {Association} {Test}:
  {III}. {Meta}-analysis of predictive validity\BBCQ\
\newblock {\Bem Journal of Personality and Social Psychology}, {\Bem 97\/}(1),
  17--41.
\newblock Place: US Publisher: American Psychological Association.

\bibitem[\protect\BCAY{Hambleton\ \BBA\ Swaminathan}{Hambleton\ \BBA\
  Swaminathan}{2013}]{hambleton2013IRT}
Hambleton, R.~K.\BBACOMMA\  \BBA\ Swaminathan, H. \BBOP2013\BBCP.
\newblock {\Bem Item response theory: Principles and applications}.
\newblock Springer Science \& Business Media.

\bibitem[\protect\BCAY{Harrington}{Harrington}{2009}]{harrington_confirmatory_2009}
Harrington, D. \BBOP2009\BBCP.
\newblock {\Bem Confirmatory {Factor} {Analysis}}.
\newblock Oxford University Press, USA.

\bibitem[\protect\BCAY{Hogenboom, Schulz,\ \BBA\ Van~Maanen}{Hogenboom
  et~al.}{2023}]{hogenboominprep}
Hogenboom, S. A.~M., Schulz, K., \BBA\ Van~Maanen, L. \BBOP2023\BBCP.
\newblock \BBOQ Implicit association tests: Stimuli validation from participant
  responses\BBCQ\
\newblock Forthcoming.

\bibitem[\protect\BCAY{Hooker, Moorosi, Clark, Bengio,\ \BBA\ Denton}{Hooker
  et~al.}{2020}]{hooker2020characterising}
Hooker, S., Moorosi, N., Clark, G., Bengio, S., \BBA\ Denton, E.
  \BBOP2020\BBCP.
\newblock \BBOQ Characterising bias in compressed models\BBCQ\
\newblock {\Bem arXiv preprint arXiv:2010.03058}.

\bibitem[\protect\BCAY{Hutchinson, Prabhakaran, Denton, Webster, Zhong,\ \BBA\
  Denuyl}{Hutchinson et~al.}{2020}]{hutchinson-etal-2020-social}
Hutchinson, B., Prabhakaran, V., Denton, E., Webster, K., Zhong, Y., \BBA\
  Denuyl, S. \BBOP2020\BBCP.
\newblock \BBOQ Social biases in {NLP} models as barriers for persons with
  disabilities\BBCQ\
\newblock In {\Bem Proceedings of the 58th Annual Meeting of the Association
  for Computational Linguistics}, \BPGS\ 5491--5501, Online. Association for
  Computational Linguistics.

\bibitem[\protect\BCAY{Jacobs\ \BBA\ Wallach}{Jacobs\ \BBA\
  Wallach}{2021}]{jacobs_measurement_2021}
Jacobs, A.~Z.\BBACOMMA\  \BBA\ Wallach, H. \BBOP2021\BBCP.
\newblock \BBOQ Measurement and {Fairness}\BBCQ\
\newblock In {\Bem Proceedings of the 2021 {ACM} {Conference} on {Fairness},
  {Accountability}, and {Transparency}}, {FAccT} '21, \BPGS\ 375--385, New
  York, NY, USA. Association for Computing Machinery.

\bibitem[\protect\BCAY{Jiao\ \BBA\ Luo}{Jiao\ \BBA\
  Luo}{2021}]{jiao_gender_2021}
Jiao, M.\BBACOMMA\  \BBA\ Luo, Z. \BBOP2021\BBCP.
\newblock \BBOQ Gender {Bias} {Hidden} {Behind} {Chinese} {Word} {Embeddings}:
  {The} {Case} of {Chinese} {Adjectives}\BBCQ\
\newblock In {\Bem Proceedings of the 3rd {Workshop} on {Gender} {Bias} in
  {Natural} {Language} {Processing}}, \BPGS\ 8--15, Online. Association for
  Computational Linguistics.

\bibitem[\protect\BCAY{Kidd\ \BBA\ Birhane}{Kidd\ \BBA\
  Birhane}{2023}]{kidd2023ai}
Kidd, C.\BBACOMMA\  \BBA\ Birhane, A. \BBOP2023\BBCP.
\newblock \BBOQ How {AI} can distort human beliefs\BBCQ\
\newblock {\Bem Science}, {\Bem 380\/}(6651), 1222--1223.

\bibitem[\protect\BCAY{Kiritchenko, Nejadgholi,\ \BBA\ Fraser}{Kiritchenko
  et~al.}{2021}]{kiritchenko_confronting_2021}
Kiritchenko, S., Nejadgholi, I., \BBA\ Fraser, K.~C. \BBOP2021\BBCP.
\newblock \BBOQ Confronting {Abusive} {Language} {Online}: {A} {Survey} from
  the {Ethical} and {Human} {Rights} {Perspective}\BBCQ\
\newblock {\Bem Journal of Artificial Intelligence Research}, {\Bem 71},
  431--478.

\bibitem[\protect\BCAY{Kline}{Kline}{2014}]{kline2014FactorAnalysis}
Kline, P. \BBOP2014\BBCP.
\newblock {\Bem An easy guide to factor analysis}.
\newblock Routledge.

\bibitem[\protect\BCAY{Krumpal}{Krumpal}{2013}]{krumpal2013Socialdesirability}
Krumpal, I. \BBOP2013\BBCP.
\newblock \BBOQ Determinants of social desirability bias in sensitive surveys:
  a literature review\BBCQ\
\newblock {\Bem Quality \& quantity}, {\Bem 47\/}(4), 2025--2047.

\bibitem[\protect\BCAY{Lalor, Wu,\ \BBA\ Yu}{Lalor
  et~al.}{2016}]{lalor2016building}
Lalor, J.~P., Wu, H., \BBA\ Yu, H. \BBOP2016\BBCP.
\newblock \BBOQ Building an evaluation scale using item response theory\BBCQ\
\newblock In {\Bem Proceedings of the Conference on Empirical Methods in
  Natural Language Processing. Conference on Empirical Methods in Natural
  Language Processing}, \lowercase{\BVOL}\ 2016, \BPG\ 648. NIH Public Access.

\bibitem[\protect\BCAY{Lam, Arnott, Beale, Chandramouli, Hilfiker-Kleiner,
  Kaye, Ky, Santema, Sliwa,\ \BBA\ Voors}{Lam et~al.}{2019}]{lam_sex_2019}
Lam, C. S.~P., Arnott, C., Beale, A.~L., Chandramouli, C., Hilfiker-Kleiner,
  D., Kaye, D.~M., Ky, B., Santema, B.~T., Sliwa, K., \BBA\ Voors, A.~A.
  \BBOP2019\BBCP.
\newblock \BBOQ Sex differences in heart failure\BBCQ\
\newblock {\Bem European Heart Journal}, {\Bem 40\/}(47), 3859--3868c.

\bibitem[\protect\BCAY{Laskar, Bari, Rahman, Bhuiyan, Joty,\ \BBA\
  Huang}{Laskar et~al.}{2023}]{laskar2023systematic}
Laskar, M. T.~R., Bari, M.~S., Rahman, M., Bhuiyan, M. A.~H., Joty, S., \BBA\
  Huang, J.~X. \BBOP2023\BBCP.
\newblock \BBOQ A systematic study and comprehensive evaluation of chatgpt on
  benchmark datasets\BBCQ\
\newblock {\Bem arXiv preprint arXiv:2305.18486}.

\bibitem[\protect\BCAY{Limisiewicz\ \BBA\ Mareček}{Limisiewicz\ \BBA\
  Mareček}{2022}]{limisiewicz_dont_2022}
Limisiewicz, T.\BBACOMMA\  \BBA\ Mareček, D. \BBOP2022\BBCP.
\newblock \BBOQ Don't {Forget} {About} {Pronouns}: {Removing} {Gender} {Bias}
  in {Language} {Models} {Without} {Losing} {Factual} {Gender}
  {Information}\BBCQ\
\newblock In {\Bem Proceedings of the 4th {Workshop} on {Gender} {Bias} in
  {Natural} {Language} {Processing} ({GeBNLP})}, \BPGS\ 17--29, Seattle,
  Washington. Association for Computational Linguistics.

\bibitem[\protect\BCAY{Liu, Yuan, Fu, Jiang, Hayashi,\ \BBA\ Neubig}{Liu
  et~al.}{2023}]{liu_pre-train_2023}
Liu, P., Yuan, W., Fu, J., Jiang, Z., Hayashi, H., \BBA\ Neubig, G.
  \BBOP2023\BBCP.
\newblock \BBOQ Pre-train, {Prompt}, and {Predict}: {A} {Systematic} {Survey}
  of {Prompting} {Methods} in {Natural} {Language} {Processing}\BBCQ\
\newblock {\Bem ACM Computing Surveys}, {\Bem 55\/}(9), 195:1--195:35.

\bibitem[\protect\BCAY{Longpre, Hou, Vu, Webson, Chung, Tay, Zhou, Le, Zoph,
  Wei,\ \BBA\ Roberts}{Longpre et~al.}{2023}]{longpre_flan_2023}
Longpre, S., Hou, L., Vu, T., Webson, A., Chung, H.~W., Tay, Y., Zhou, D., Le,
  Q.~V., Zoph, B., Wei, J., \BBA\ Roberts, A. \BBOP2023\BBCP.
\newblock \BBOQ The {Flan} collection: Designing data and methods for effective
  instruction tuning\BBCQ\
\newblock In {\Bem International Conference on Machine Learning, {ICML} 2023,
  23-29 July 2023, Honolulu, Hawaii, {USA}}, \lowercase{\BVOL}\ 202 of {\Bem
  Proceedings of Machine Learning Research}, \BPGS\ 22631--22648. {PMLR}.

\bibitem[\protect\BCAY{Malik, Dev, Nishi, Peng,\ \BBA\ Chang}{Malik
  et~al.}{2022}]{malik_socially_2022}
Malik, V., Dev, S., Nishi, A., Peng, N., \BBA\ Chang, K.-W. \BBOP2022\BBCP.
\newblock \BBOQ Socially {Aware} {Bias} {Measurements} for {Hindi} {Language}
  {Representations}\BBCQ\
\newblock In {\Bem Proceedings of the 2022 {Conference} of the {North}
  {American} {Chapter} of the {Association} for {Computational} {Linguistics}:
  {Human} {Language} {Technologies}}, \BPGS\ 1041--1052, Seattle, United
  States. Association for Computational Linguistics.

\bibitem[\protect\BCAY{Mathet, Widlöcher,\ \BBA\ Métivier}{Mathet
  et~al.}{2015}]{mathet_unified_2015}
Mathet, Y., Widlöcher, A., \BBA\ Métivier, J.-P. \BBOP2015\BBCP.
\newblock \BBOQ The {Unified} and {Holistic} {Method} {Gamma} ($\gamma$) for
  {Inter}-{Annotator} {Agreement} {Measure} and {Alignment}\BBCQ\
\newblock {\Bem Computational Linguistics}, {\Bem 41\/}(3), 437--479.

\bibitem[\protect\BCAY{May, Wang, Bordia, Bowman,\ \BBA\ Rudinger}{May
  et~al.}{2019}]{may_measuring_2019}
May, C., Wang, A., Bordia, S., Bowman, S.~R., \BBA\ Rudinger, R.
  \BBOP2019\BBCP.
\newblock \BBOQ On {Measuring} {Social} {Biases} in {Sentence} {Encoders}\BBCQ\
\newblock In {\Bem Proceedings of the 2019 {Conference} of the {North}
  {American} {Chapter} of the {Association} for {Computational} {Linguistics}:
  {Human} {Language} {Technologies}, {Volume} 1 ({Long} and {Short} {Papers})},
  \BPGS\ 622--628, Minneapolis, Minnesota. Association for Computational
  Linguistics.

\bibitem[\protect\BCAY{Meade, Poole-Dayan,\ \BBA\ Reddy}{Meade
  et~al.}{2022}]{meade_empirical_2022}
Meade, N., Poole-Dayan, E., \BBA\ Reddy, S. \BBOP2022\BBCP.
\newblock \BBOQ An {Empirical} {Survey} of the {Effectiveness} of {Debiasing}
  {Techniques} for {Pre}-trained {Language} {Models}\BBCQ\
\newblock In {\Bem Proceedings of the 60th {Annual} {Meeting} of the
  {Association} for {Computational} {Linguistics} ({Volume} 1: {Long}
  {Papers})}, \BPGS\ 1878--1898, Dublin, Ireland. Association for Computational
  Linguistics.

\bibitem[\protect\BCAY{Montuschi, Gatteschi, Lamberti, Sanna,\ \BBA\
  Demartini}{Montuschi et~al.}{2013}]{montuschi2013job}
Montuschi, P., Gatteschi, V., Lamberti, F., Sanna, A., \BBA\ Demartini, C.
  \BBOP2013\BBCP.
\newblock \BBOQ Job recruitment and job seeking processes: how technology can
  help\BBCQ\
\newblock {\Bem It professional}, {\Bem 16\/}(5), 41--49.

\bibitem[\protect\BCAY{Nadeem, Bethke,\ \BBA\ Reddy}{Nadeem
  et~al.}{2021}]{nadeem_stereoset_2021}
Nadeem, M., Bethke, A., \BBA\ Reddy, S. \BBOP2021\BBCP.
\newblock \BBOQ {StereoSet}: {Measuring} stereotypical bias in pretrained
  language models\BBCQ\
\newblock In {\Bem Proceedings of the 59th {Annual} {Meeting} of the
  {Association} for {Computational} {Linguistics} and the 11th {International}
  {Joint} {Conference} on {Natural} {Language} {Processing} ({Volume} 1: {Long}
  {Papers})}, \BPGS\ 5356--5371, Online. Association for Computational
  Linguistics.

\bibitem[\protect\BCAY{Nangia, Vania, Bhalerao,\ \BBA\ Bowman}{Nangia
  et~al.}{2020}]{nangia_crows-pairs_2020}
Nangia, N., Vania, C., Bhalerao, R., \BBA\ Bowman, S.~R. \BBOP2020\BBCP.
\newblock \BBOQ {CrowS}-{Pairs}: {A} {Challenge} {Dataset} for {Measuring}
  {Social} {Biases} in {Masked} {Language} {Models}\BBCQ\
\newblock In {\Bem Proceedings of the 2020 {Conference} on {Empirical}
  {Methods} in {Natural} {Language} {Processing} ({EMNLP})}, \BPGS\ 1953--1967,
  Online. Association for Computational Linguistics.

\bibitem[\protect\BCAY{Newton\ \BBA\ Shaw}{Newton\ \BBA\
  Shaw}{2013}]{newton_standards_2013}
Newton, P.~E.\BBACOMMA\  \BBA\ Shaw, S.~D. \BBOP2013\BBCP.
\newblock \BBOQ Standards for talking and thinking about validity\BBCQ\
\newblock {\Bem Psychological Methods}, {\Bem 18}, 301--319.
\newblock Place: US Publisher: American Psychological Association.

\bibitem[\protect\BCAY{Nissim, van Noord,\ \BBA\ van~der Goot}{Nissim
  et~al.}{2020}]{nissim2020fair}
Nissim, M., van Noord, R., \BBA\ van~der Goot, R. \BBOP2020\BBCP.
\newblock \BBOQ Fair is better than sensational: Man is to doctor as woman is
  to doctor\BBCQ\
\newblock {\Bem Computational Linguistics}, {\Bem 46\/}(2), 487--497.

\bibitem[\protect\BCAY{Nosek, Alter, Banks, Borsboom, Bowman, Breckler, Buck,
  Chambers, Chin, Christensen, Contestabile, Dafoe, Eich, Freese, Glennerster,
  Goroff, Green, Hesse, Humphreys, Ishiyama, Karlan, Kraut, Lupia, Mabry,
  Madon, Malhotra, Mayo-Wilson, McNutt, Miguel, Paluck, Simonsohn, Soderberg,
  Spellman, Turitto, VandenBos, Vazire, Wagenmakers, Wilson,\ \BBA\
  Yarkoni}{Nosek et~al.}{2015}]{nosek_promoting_2015}
Nosek, B.~A., Alter, G., Banks, G.~C., Borsboom, D., Bowman, S.~D., Breckler,
  S.~J., Buck, S., Chambers, C.~D., Chin, G., Christensen, G., Contestabile,
  M., Dafoe, A., Eich, E., Freese, J., Glennerster, R., Goroff, D., Green,
  D.~P., Hesse, B., Humphreys, M., Ishiyama, J., Karlan, D., Kraut, A., Lupia,
  A., Mabry, P., Madon, T., Malhotra, N., Mayo-Wilson, E., McNutt, M., Miguel,
  E., Paluck, E.~L., Simonsohn, U., Soderberg, C., Spellman, B.~A., Turitto,
  J., VandenBos, G., Vazire, S., Wagenmakers, E.~J., Wilson, R., \BBA\ Yarkoni,
  T. \BBOP2015\BBCP.
\newblock \BBOQ Promoting an open research culture\BBCQ\
\newblock {\Bem Science}, {\Bem 348\/}(6242), 1422--1425.
\newblock Publisher: American Association for the Advancement of Science.

\bibitem[\protect\BCAY{Névéol, Dupont, Bezançon,\ \BBA\ Fort}{Névéol
  et~al.}{2022}]{neveol_french_2022}
Névéol, A., Dupont, Y., Bezançon, J., \BBA\ Fort, K. \BBOP2022\BBCP.
\newblock \BBOQ French {CrowS}-{Pairs}: {Extending} a challenge dataset for
  measuring social bias in masked language models to a language other than
  {English}\BBCQ\
\newblock In {\Bem Proceedings of the 60th {Annual} {Meeting} of the
  {Association} for {Computational} {Linguistics} ({Volume} 1: {Long}
  {Papers})}, \BPGS\ 8521--8531, Dublin, Ireland. Association for Computational
  Linguistics.

\bibitem[\protect\BCAY{Orgad\ \BBA\ Belinkov}{Orgad\ \BBA\
  Belinkov}{2022}]{orgad2022choose}
Orgad, H.\BBACOMMA\  \BBA\ Belinkov, Y. \BBOP2022\BBCP.
\newblock \BBOQ Choose your lenses: Flaws in gender bias evaluation\BBCQ\
\newblock {\Bem GeBNLP 2022}, 151.

\bibitem[\protect\BCAY{Prystawski, Grant, Nematzadeh, Lee, Stevenson,\ \BBA\
  Xu}{Prystawski et~al.}{2022}]{prystawski_emergence_2022}
Prystawski, B., Grant, E., Nematzadeh, A., Lee, S. W.~S., Stevenson, S., \BBA\
  Xu, Y. \BBOP2022\BBCP.
\newblock \BBOQ The {Emergence} of {Gender} {Associations} in {Child}
  {Language} {Development}\BBCQ\
\newblock {\Bem Cognitive Science}, {\Bem 46\/}(6), e13146.

\bibitem[\protect\BCAY{Sanh, Webson, Raffel, Bach, Sutawika, Alyafeai, Chaffin,
  Stiegler, Le~Scao, Raja, et~al.}{Sanh et~al.}{2022}]{sanh2022multitask}
Sanh, V., Webson, A., Raffel, C., Bach, S.~H., Sutawika, L., Alyafeai, Z.,
  Chaffin, A., Stiegler, A., Le~Scao, T., Raja, A., et~al. \BBOP2022\BBCP.
\newblock \BBOQ Multitask prompted training enables zero-shot task
  generalization\BBCQ\
\newblock In {\Bem ICLR 2022-Tenth International Conference on Learning
  Representations}.

\bibitem[\protect\BCAY{Scao, Fan, Akiki, Pavlick, Ili{\'c}, Hesslow,
  Castagn{\'e}, Luccioni, Yvon, Gall{\'e}, et~al.}{Scao
  et~al.}{2022}]{scao_bloom_2022b}
Scao, T.~L., Fan, A., Akiki, C., Pavlick, E., Ili{\'c}, S., Hesslow, D.,
  Castagn{\'e}, R., Luccioni, A.~S., Yvon, F., Gall{\'e}, M., et~al.
  \BBOP2022\BBCP.
\newblock \BBOQ Bloom: A 176b-parameter open-access multilingual language
  model\BBCQ\
\newblock {\Bem arXiv preprint arXiv:2211.05100}.

\bibitem[\protect\BCAY{Schick, Udupa,\ \BBA\ Sch{\"u}tze}{Schick
  et~al.}{2021}]{schick2021self}
Schick, T., Udupa, S., \BBA\ Sch{\"u}tze, H. \BBOP2021\BBCP.
\newblock \BBOQ Self-diagnosis and self-debiasing: A proposal for reducing
  corpus-based bias in nlp\BBCQ\
\newblock {\Bem Transactions of the Association for Computational Linguistics},
  {\Bem 9}, 1408--1424.

\bibitem[\protect\BCAY{Sedoc\ \BBA\ Ungar}{Sedoc\ \BBA\
  Ungar}{2019}]{sedoc_role_2019}
Sedoc, J.\BBACOMMA\  \BBA\ Ungar, L. \BBOP2019\BBCP.
\newblock \BBOQ The {Role} of {Protected} {Class} {Word} {Lists} in {Bias}
  {Identification} of {Contextualized} {Word} {Representations}\BBCQ\
\newblock In {\Bem Proceedings of the {First} {Workshop} on {Gender} {Bias} in
  {Natural} {Language} {Processing}}, \BPGS\ 55--61, Florence, Italy.
  Association for Computational Linguistics.

\bibitem[\protect\BCAY{Seshadri, Pezeshkpour,\ \BBA\ Singh}{Seshadri
  et~al.}{2022}]{seshadri_quantifying_2022}
Seshadri, P., Pezeshkpour, P., \BBA\ Singh, S. \BBOP2022\BBCP.
\newblock \BBOQ Quantifying social biases using templates is unreliable\BBCQ\
\newblock {\Bem arXiv preprint arXiv:2210.04337}.

\bibitem[\protect\BCAY{Stanczak\ \BBA\ Augenstein}{Stanczak\ \BBA\
  Augenstein}{2021}]{stanczak_survey_2021}
Stanczak, K.\BBACOMMA\  \BBA\ Augenstein, I. \BBOP2021\BBCP.
\newblock \BBOQ A survey on gender bias in natural language processing\BBCQ\
\newblock {\Bem arXiv preprint arXiv:2112.14168}.

\bibitem[\protect\BCAY{Stanovsky, Smith,\ \BBA\ Zettlemoyer}{Stanovsky
  et~al.}{2019}]{stanovsky2019evaluating}
Stanovsky, G., Smith, N.~A., \BBA\ Zettlemoyer, L. \BBOP2019\BBCP.
\newblock \BBOQ Evaluating gender bias in machine translation\BBCQ\
\newblock In {\Bem Proceedings of the 57th Annual Meeting of the Association
  for Computational Linguistics}, \BPGS\ 1679--1684.

\bibitem[\protect\BCAY{Talat, Névéol, Biderman, Clinciu, Dey, Longpre,
  Luccioni, Masoud, Mitchell, Radev, Sharma, Subramonian, Tae, Tan,
  Tunuguntla,\ \BBA\ van~der Wal}{Talat et~al.}{2022}]{talat_you_2022}
Talat, Z., Névéol, A., Biderman, S., Clinciu, M., Dey, M., Longpre, S.,
  Luccioni, S., Masoud, M., Mitchell, M., Radev, D., Sharma, S., Subramonian,
  A., Tae, J., Tan, S., Tunuguntla, D., \BBA\ van~der Wal, O. \BBOP2022\BBCP.
\newblock \BBOQ You reap what you sow: {On} the {Challenges} of {Bias}
  {Evaluation} {Under} {Multilingual} {Settings}\BBCQ\
\newblock In {\Bem Proceedings of {BigScience} {Episode} \#5 – {Workshop} on
  {Challenges} \& {Perspectives} in {Creating} {Large} {Language} {Models}},
  \BPGS\ 26--41, virtual+Dublin. Association for Computational Linguistics.

\bibitem[\protect\BCAY{Tontodimamma, Fontanella, Anzani,\ \BBA\
  Basile}{Tontodimamma et~al.}{2022}]{tontodimamma_italian_2022}
Tontodimamma, A., Fontanella, L., Anzani, S., \BBA\ Basile, V. \BBOP2022\BBCP.
\newblock \BBOQ An {Italian} lexical resource for incivility detection in
  online discourses\BBCQ\
\newblock {\Bem Quality \& Quantity}, {\Bem 56}.

\bibitem[\protect\BCAY{Verma, Vieweg, Corvey, Palen, Martin, Palmer, Schram,\
  \BBA\ Anderson}{Verma et~al.}{2011}]{verma_natural_2011}
Verma, S., Vieweg, S., Corvey, W., Palen, L., Martin, J., Palmer, M., Schram,
  A., \BBA\ Anderson, K. \BBOP2011\BBCP.
\newblock \BBOQ Natural {Language} {Processing} to the {Rescue}? {Extracting}
  "{Situational} {Awareness}" {Tweets} {During} {Mass} {Emergency}\BBCQ\
\newblock {\Bem Proceedings of the International AAAI Conference on Web and
  Social Media}, {\Bem 5\/}(1), 385--392.
\newblock Number: 1.

\bibitem[\protect\BCAY{Vig, Gehrmann, Belinkov, Qian, Nevo, Singer,\ \BBA\
  Shieber}{Vig et~al.}{2020}]{vig_investigating_2020}
Vig, J., Gehrmann, S., Belinkov, Y., Qian, S., Nevo, D., Singer, Y., \BBA\
  Shieber, S. \BBOP2020\BBCP.
\newblock \BBOQ Investigating {Gender} {Bias} in {Language} {Models} {Using}
  {Causal} {Mediation} {Analysis}\BBCQ\
\newblock In {\Bem Advances in {Neural} {Information} {Processing} {Systems}},
  \lowercase{\BVOL}~33, \BPGS\ 12388--12401. Curran Associates, Inc.

\bibitem[\protect\BCAY{Wagner, Graells-Garrido, Garcia,\ \BBA\ Menczer}{Wagner
  et~al.}{2016}]{wagner_women_2016}
Wagner, C., Graells-Garrido, E., Garcia, D., \BBA\ Menczer, F. \BBOP2016\BBCP.
\newblock \BBOQ Women through the glass ceiling: gender asymmetries in
  {Wikipedia}\BBCQ\
\newblock {\Bem EPJ Data Science}, {\Bem 5\/}(1), 5.

\bibitem[\protect\BCAY{Walter, Kirschner, Eger, Glavaš, Lauscher,\ \BBA\
  Ponzetto}{Walter et~al.}{2021}]{walter_diachronic_2021}
Walter, T., Kirschner, C., Eger, S., Glavaš, G., Lauscher, A., \BBA\ Ponzetto,
  S.~P. \BBOP2021\BBCP.
\newblock \BBOQ Diachronic {Analysis} of {German} {Parliamentary}
  {Proceedings}: {Ideological} {Shifts} through the {Lens} of {Political}
  {Biases}\BBCQ\
\newblock In {\Bem 2021 {ACM}/{IEEE} {Joint} {Conference} on {Digital}
  {Libraries} ({JCDL})}, \BPGS\ 51--60.

\bibitem[\protect\BCAY{{\VAN{Wal}{Van der}{van der}}~Wal, Jumelet, Schulz,\
  \BBA\ Zuidema}{{\VAN{Wal}{Van der}{van der}}~Wal
  et~al.}{2022}]{van_der_wal_birth_2022b}
{\VAN{Wal}{Van der}{van der}}~Wal, O., Jumelet, J., Schulz, K., \BBA\ Zuidema,
  W. \BBOP2022\BBCP.
\newblock \BBOQ The {Birth} of {Bias}: {A} case study on the evolution of
  gender bias in an {English} language model\BBCQ\
\newblock In {\Bem Proceedings of the 4th {Workshop} on {Gender} {Bias} in
  {Natural} {Language} {Processing} ({GeBNLP})}, \BPGS\ 75--75, Seattle,
  Washington. Association for Computational Linguistics.

\bibitem[\protect\BCAY{Wang, Wang, Rastegar-Mojarad, Moon, Shen, Afzal, Liu,
  Zeng, Mehrabi, Sohn, et~al.}{Wang et~al.}{2018}]{wang2018clinical}
Wang, Y., Wang, L., Rastegar-Mojarad, M., Moon, S., Shen, F., Afzal, N., Liu,
  S., Zeng, Y., Mehrabi, S., Sohn, S., et~al. \BBOP2018\BBCP.
\newblock \BBOQ Clinical information extraction applications: a literature
  review\BBCQ\
\newblock {\Bem Journal of biomedical informatics}, {\Bem 77}, 34--49.

\bibitem[\protect\BCAY{Warrens}{Warrens}{2015}]{warrens2015cronbach}
Warrens, M.~J. \BBOP2015\BBCP.
\newblock \BBOQ On cronbach’s alpha as the mean of all split-half
  reliabilities\BBCQ\
\newblock In {\Bem Quantitative Psychology Research: The 78th Annual Meeting of
  the Psychometric Society}, \BPGS\ 293--300. Springer.

\bibitem[\protect\BCAY{Way}{Way}{2018}]{way2018quality}
Way, A. \BBOP2018\BBCP.
\newblock \BBOQ Quality expectations of machine translation\BBCQ\
\newblock In {\Bem Translation quality assessment}, \BPGS\ 159--178. Springer.

\bibitem[\protect\BCAY{Webster, Wang, Tenney, Beutel, Pitler, Pavlick, Chen,
  Chi,\ \BBA\ Petrov}{Webster et~al.}{2020}]{webster2020measuring}
Webster, K., Wang, X., Tenney, I., Beutel, A., Pitler, E., Pavlick, E., Chen,
  J., Chi, E., \BBA\ Petrov, S. \BBOP2020\BBCP.
\newblock \BBOQ Measuring and reducing gendered correlations in pre-trained
  models\BBCQ\
\newblock {\Bem arXiv preprint arXiv:2010.06032}.

\bibitem[\protect\BCAY{Weinberg}{Weinberg}{2022}]{weinberg_rethinking_2022}
Weinberg, L. \BBOP2022\BBCP.
\newblock \BBOQ Rethinking {Fairness}: {An} {Interdisciplinary} {Survey} of
  {Critiques} of {Hegemonic} {ML} {Fairness} {Approaches}\BBCQ\
\newblock {\Bem Journal of Artificial Intelligence Research}, {\Bem 74},
  75--109.

\bibitem[\protect\BCAY{Whitlock\ \BBA\ Schluter}{Whitlock\ \BBA\
  Schluter}{2015}]{whitlock2015analysis}
Whitlock, M.\BBACOMMA\  \BBA\ Schluter, D. \BBOP2015\BBCP.
\newblock {\Bem The analysis of biological data}, \lowercase{\BVOL}\ 768.
\newblock Roberts Publishers.

\bibitem[\protect\BCAY{Winham, de~Andrade,\ \BBA\ Miller}{Winham
  et~al.}{2015}]{winham_genetics_2015}
Winham, S.~J., de~Andrade, M., \BBA\ Miller, V.~M. \BBOP2015\BBCP.
\newblock \BBOQ Genetics of cardiovascular disease: {Importance} of sex and
  ethnicity\BBCQ\
\newblock {\Bem Atherosclerosis}, {\Bem 241\/}(1), 219--228.

\bibitem[\protect\BCAY{Wong\ \BBA\ Paritosh}{Wong\ \BBA\
  Paritosh}{2022}]{wong2022Interrater}
Wong, K.\BBACOMMA\  \BBA\ Paritosh, P. \BBOP2022\BBCP.
\newblock \BBOQ k-rater reliability: The correct unit of reliability for
  aggregated human annotations\BBCQ\
\newblock In {\Bem Proceedings of the 60th Annual Meeting of the Association
  for Computational Linguistics (Volume 2: Short Papers)}, \BPGS\ 378--384.

\bibitem[\protect\BCAY{Wong, Paritosh,\ \BBA\ Aroyo}{Wong
  et~al.}{2021}]{wong_cross-replication_2021}
Wong, K., Paritosh, P., \BBA\ Aroyo, L. \BBOP2021\BBCP.
\newblock \BBOQ Cross-replication {Reliability} - {An} {Empirical} {Approach}
  to {Interpreting} {Inter}-rater {Reliability}\BBCQ\
\newblock In {\Bem Proceedings of the 59th {Annual} {Meeting} of the
  {Association} for {Computational} {Linguistics} and the 11th {International}
  {Joint} {Conference} on {Natural} {Language} {Processing} ({Volume} 1: {Long}
  {Papers})}, \BPGS\ 7053--7065, Online. Association for Computational
  Linguistics.

\bibitem[\protect\BCAY{Zeinert, Inie,\ \BBA\ Derczynski}{Zeinert
  et~al.}{2021}]{zeinert_annotating_2021}
Zeinert, P., Inie, N., \BBA\ Derczynski, L. \BBOP2021\BBCP.
\newblock \BBOQ Annotating {Online} {Misogyny}\BBCQ\
\newblock In {\Bem Proceedings of the 59th {Annual} {Meeting} of the
  {Association} for {Computational} {Linguistics} and the 11th {International}
  {Joint} {Conference} on {Natural} {Language} {Processing} ({Volume} 1: {Long}
  {Papers})}, \BPGS\ 3181--3197, Online. Association for Computational
  Linguistics.

\bibitem[\protect\BCAY{Zhang, Sneyd,\ \BBA\ Stevenson}{Zhang
  et~al.}{2020}]{zhang_robustness_2020}
Zhang, H., Sneyd, A., \BBA\ Stevenson, M. \BBOP2020\BBCP.
\newblock \BBOQ Robustness and {Reliability} of {Gender} {Bias} {Assessment} in
  {Word} {Embeddings}: {The} {Role} of {Base} {Pairs}\BBCQ\
\newblock In {\Bem Proceedings of the 1st {Conference} of the {Asia}-{Pacific}
  {Chapter} of the {Association} for {Computational} {Linguistics} and the 10th
  {International} {Joint} {Conference} on {Natural} {Language} {Processing}},
  \BPGS\ 759--769, Suzhou, China. Association for Computational Linguistics.

\bibitem[\protect\BCAY{Zhang, Roller, Goyal, Artetxe, Chen, Chen, Dewan, Diab,
  Li, Lin, et~al.}{Zhang et~al.}{2022a}]{zhang_opt_2022}
Zhang, S., Roller, S., Goyal, N., Artetxe, M., Chen, M., Chen, S., Dewan, C.,
  Diab, M., Li, X., Lin, X.~V., et~al. \BBOP2022a\BBCP.
\newblock \BBOQ Opt: Open pre-trained transformer language models\BBCQ\
\newblock {\Bem arXiv preprint arXiv:2205.01068}.

\bibitem[\protect\BCAY{Zhang, Zhang, Halpern, Patel,\ \BBA\ Scharenborg}{Zhang
  et~al.}{2022b}]{zhang2022mitigating}
Zhang, Y., Zhang, Y., Halpern, B.~M., Patel, T., \BBA\ Scharenborg, O.
  \BBOP2022b\BBCP.
\newblock \BBOQ Mitigating bias against non-native accents\BBCQ\
\newblock In {\Bem Proceedings of the Annual Conference of the International
  Speech Communication Association, INTERSPEECH}, \lowercase{\BVOL}\ 2022,
  \BPGS\ 3168--3172.

\bibitem[\protect\BCAY{Zhao, Wang, Yatskar, Cotterell, Ordonez,\ \BBA\
  Chang}{Zhao et~al.}{2019}]{zhao_gender_2019}
Zhao, J., Wang, T., Yatskar, M., Cotterell, R., Ordonez, V., \BBA\ Chang, K.-W.
  \BBOP2019\BBCP.
\newblock \BBOQ Gender {Bias} in {Contextualized} {Word} {Embeddings}\BBCQ\
\newblock In {\Bem Proceedings of the 2019 {Conference} of the {North}
  {American} {Chapter} of the {Association} for {Computational} {Linguistics}:
  {Human} {Language} {Technologies}, {Volume} 1 ({Long} and {Short} {Papers})},
  \BPGS\ 629--634, Minneapolis, Minnesota. Association for Computational
  Linguistics.

\bibitem[\protect\BCAY{Zhao, Wang, Yatskar, Ordonez,\ \BBA\ Chang}{Zhao
  et~al.}{2018}]{zhao_gender_2018}
Zhao, J., Wang, T., Yatskar, M., Ordonez, V., \BBA\ Chang, K.-W.
  \BBOP2018\BBCP.
\newblock \BBOQ Gender {Bias} in {Coreference} {Resolution}: {Evaluation} and
  {Debiasing} {Methods}\BBCQ\
\newblock In {\Bem Proceedings of the 2018 {Conference} of the {North}
  {American} {Chapter} of the {Association} for {Computational} {Linguistics}:
  {Human} {Language} {Technologies}, {Volume} 2 ({Short} {Papers})}, \BPGS\
  15--20, New Orleans, Louisiana. Association for Computational Linguistics.

\bibitem[\protect\BCAY{Zhong, Xiao, Tu, Zhang, Liu,\ \BBA\ Sun}{Zhong
  et~al.}{2020}]{zhong2020does}
Zhong, H., Xiao, C., Tu, C., Zhang, T., Liu, Z., \BBA\ Sun, M. \BBOP2020\BBCP.
\newblock \BBOQ How does nlp benefit legal system: A summary of legal
  artificial intelligence\BBCQ\
\newblock In {\Bem Proceedings of the 58th Annual Meeting of the Association
  for Computational Linguistics}, \BPGS\ 5218--5230.

\end{thebibliography}
\bibliographystyle{theapa}

\end{document}